\documentclass[conference]{IEEEtran}
\IEEEoverridecommandlockouts

\usepackage{cite}
\usepackage{amsmath,amssymb,amsfonts}
\usepackage{algorithmic}
\usepackage{graphicx}
\usepackage{textcomp}
\usepackage{xcolor}
\def\BibTeX{{\rm B\kern-.05em{\sc i\kern-.025em b}\kern-.08em
    T\kern-.1667em\lower.7ex\hbox{E}\kern-.125emX}}

\usepackage{mathtools} 
\usepackage{booktabs} 
\usepackage{tikz} 

\usepackage{algorithm} 
\usepackage{amssymb}       
\usepackage{hyperref}
\usepackage{stmaryrd}      
\usepackage{xspace}
\usetikzlibrary{arrows,backgrounds,calc,chains,decorations.pathmorphing,fit,petri,positioning,shapes}

\begin{document}

\title{Probabilistic ML Verification \\via  Weighted Model Integration}



\author{\IEEEauthorblockN{1\textsuperscript{st} Paolo Morettin}
\IEEEauthorblockA{\textit{DISI} \\
\textit{University of Trento}\\
Trento, Italy \\
name.surname@unitn.it}
\and
\IEEEauthorblockN{2\textsuperscript{nd} Andrea Passerini}
\IEEEauthorblockA{\textit{DISI} \\
\textit{University of Trento}\\
Trento, Italy \\
name.surname@unitn.it}
\and
\IEEEauthorblockN{3\textsuperscript{rd} Roberto Sebastiani}
\IEEEauthorblockA{\textit{DISI} \\
\textit{University of Trento}\\
Trento, Italy \\
name.surname@unitn.it}}


\newcommand{\paolo}[1]{\textcolor{blue!66!green}{\textbf{Paolo}: #1}}
\newcommand{\andrea}[1]{\textcolor{red}{\textbf{Andrea}: #1}}
\newcommand{\roberto}[1]{\textcolor{purple}{\textbf{Roberto}: #1}}

\newcommand{\customparagraph}[1]{\noindent\textbf{#1}}
\newcommand{\mstd}[2]{#1${\scriptstyle (\pm#2)}$}

\newcommand{\set}[1]{\ensuremath{\{{#1}\}}\xspace}
\newcommand{\defas}{\ensuremath{\stackrel{\text{\tiny def}}{=}}\xspace}
\newcommand{\alg}[1]{\mathsf{#1}}

\newcommand{\prob}[1]{P(#1)\xspace}
\newcommand{\cprob}[2]{\prob{#1 \:|\: #2}\xspace}

\newcommand{\data}{\mathcal{D}}
\newcommand{\sys}{\mathcal{S}}
\newcommand{\pro}{\mathcal{R}}
\newcommand{\psys}{P_\sys}
\newcommand{\ppro}{P_\pro}
\newcommand{\pre}{I}
\newcommand{\post}{O}
\newcommand{\propre}{\pro_{\pre}}
\newcommand{\propost}{\pro_{\post}}
\newcommand{\encsys}{\Delta_{\sys}}
\newcommand{\encpre}{\Delta_{\pre}}
\newcommand{\encpost}{\Delta_{\post}}

\newcommand{\bX}{\mathbf{x}}
\newcommand{\bY}{\mathbf{y}}
\newcommand{\bW}{\mathbf{w}}
\newcommand{\bN}{\mathbf{n}}

\newcommand{\partition}{\Phi}
\newcommand{\npartitions}{N_P}

\newcommand{\peg}{\phantom{\neg}}
\newcommand{\ind}[1]{\llbracket #1 \rrbracket}
\newcommand{\lra}{\ensuremath{\mathcal{LRA}}\xspace}
\newcommand{\pmodels}{\ensuremath{\models_{\mathbb{B}}}}
\newcommand{\laratmodels}{\ensuremath{\models_{\lra}}}
\newcommand{\smtlra}{\ensuremath{\text{SMT}(\lra)}\xspace}
\newcommand{\allA}{\ensuremath{\mathbf{A}}\xspace}
\newcommand{\allx}{\ensuremath{\mathbf{x}}\xspace}
\newcommand{\TTA}{\ensuremath{\mathcal{T}\hspace{-.1cm}\mathcal{T}\hspace{-.1cm}\mathcal{A}}\xspace}
\newcommand{\mulra}{\ensuremath{\mu^{\lra}}\xspace}
\newcommand{\w}{\ensuremath{w}\xspace}
\newcommand{\wmuagen}[1]{\ensuremath{\w_{[#1]}}\xspace}
\newcommand{\wmu}{\wmuagen{\mu}}
\newcommand{\wxa}{\ensuremath{\w(\allx,\allA)}\xspace}
\newcommand{\support}{\ensuremath{\chi}\xspace}

\newcommand{\ite}{\mathsf{ITE}}
\newcommand{\maxsmt}{\mathsf{max}}
\newcommand{\scomp}{\mathsf{sc}}

\newtheorem{example}{Example}

\maketitle

\begin{abstract}
In machine learning (ML) verification, the majority of procedures are
non-quantitative and therefore cannot be used for verifying
probabilistic models, or be applied in domains where hard guarantees
are practically unachievable.
The probabilistic formal verification (PFV) of ML models is in its
infancy, with the existing approaches limited to specific ML
models, properties, or both.
This contrasts with standard formal methods techniques, whose
successful adoption in real-world scenarios is also due to their
support for a wide range of properties and diverse systems.
We propose a unifying framework for the PFV of ML systems based on
Weighted Model Integration (WMI), a relatively recent formalism for
probabilistic inference with algebraic and logical constraints.
Crucially, reducing the PFV of ML models to WMI enables the
verification of many properties of interest over a wide range of
systems, addressing multiple limitations of deterministic verification
and ad-hoc algorithms.
We substantiate the generality of the approach on prototypical tasks
involving the verification of group fairness, monotonicity, robustness
to noise, probabilistic local robustness and equivalence among
predictors.
We characterize the challenges related to the scalability of the
approach and, through our WMI-based perspective, we show how
successful scaling techniques in the ML verification literature can be
generalized beyond their original scope.

\end{abstract}


\section{Introduction and motivation}
\label{sec:intro}
The development of verification techniques for machine learning (ML)
models is nowadays considered a critical challenge for both the
scientific community~\cite{towards-verified-ai,concrete-problems-ai} and
regulatory bodies~\cite{ai-act}.
In the last three decades, model checking has made huge progress in
the verification of both hardware and software systems, guaranteeing
the deployment of safe systems in many fields~\cite{model-checking-handbook}.
A key aspect for this success was the development of algorithms that
are agnostic on the property and on the specifics of the model under
verification, thanks to unified representation formalisms such as
temporal logic and Buchi automata~\cite{model-checking-principles}.

The complexity and versatility of modern ML systems calls for a
unified approach to assess their trustworthiness, which often cannot
be captured by a single evaluation metric or formal property.
Most algorithms in the ML verification literature, however, target
specific classes of models~\cite{dfv-trees,reluplex,dfv-svm}
and/or properties~\cite{dfv-nn-rob-lp,dfv-trees-milp},
trading off generality for higher scalability in specific use
cases.

\begin{figure}
  \centering
  \includegraphics[width=0.25\textwidth]{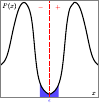}
  \caption{\label{fig:boundary} A scenario where global
    $\epsilon$-robustness is not achievable. The dashed red line and
    the blue area respectively depict the decision boundary and an
    arbitrarily small probability of the marginal $\prob{x}$.}
\end{figure}

Furthermore, most works focus on the qualitative verification of
deterministic systems, by certifying that a property is never violated
or providing a counterexample~\cite{marabou}.
This line of work has many practical use cases, such as worst-case
analysis for adversarial robustness
verification~\cite{adv-rob,dfv-trees2}, but it also has important
limitations.
First, non-quantitative approaches cannot be used for the verification
of probabilistic ML systems~\cite{prob-ml}, leaving out an important
and sizable category of use cases that include deep generative
models, Bayesian networks and probabilistic programs.
Second, these verification approaches cannot account for the
(uncertain) environment where the system is going to be deployed.
Therefore, they cannot be used to verify intrinsically probabilistic
properties, like fairness definitions based on the notion of a
population model~\cite{fairness-review}.
Moreover, they fall short when achieving hard guarantees is
impractical. For instance, deterministic global classification
robustness is an unsatisfiable criterion when even an arbitrarily
small probability mass is allocated close to the decision boundary
(Fig.~\ref{fig:boundary}).

In this work, we propose a unified perspective on the PFV of ML models
based on Weighted Model Integration
(WMI)~\cite{belle2015probabilistic}.
Broadly speaking, WMI is the task of computing probabilities of
arbitrary combinations of logical and algebraic constraints given a
structured joint distribution over both continuous and discrete
variables.
By reducing PFV tasks to WMI, our approach is both probabilistic and
property/model agnostic, supporting a wide range of use cases.
Moreover, ML models can be part of the environment definition, and
not only be a component of the system under verification, a crucial
feature when manually defining the specification is hard or
impossible.

In summary, the proposed framework satisfies three crucial desiderata
for general-purpose verification of ML systems.
\begin{itemize}  
\item[\textbf{D1})] It supports \emph{arbitrary distributions} over
  continuous and discrete domains, including complex models learned
  from data;
\item[\textbf{D2})] it supports a \emph{wide variety of deterministic
ML models}, from decision trees to neural networks and ensemble
  models, via a single representation language;
\item[\textbf{D3})] it enables the verification of a \emph{broad range
of properties} of interest.
\end{itemize}
Our experiments substantiate the generality of the approach by
considering group fairness, monotonicity, robustness to noise,
probabilistic local robustness and equivalence among predictors.
In terms of ML systems, the prototypical tasks we consider involve
verifying neural networks with rectified linear units (ReLU NNs),
random forests (RFs) or \emph{both}.
In contrast with the majority of approaches, where the system is
verified with very simple notions of environment, in our framework the
prior over the system's input can be a ML model itself. In real-world
applications, we expect these certified priors to complement
the properties as part of the requirement.
To the best of our knowledge, there is no work in the ML verification
literature providing a similarly broad perspective.

To fully harness the capabilities of this framework and apply it to
large ML systems, however, multiple scalability challenges must be
overcome.
In this work, we analyze the different factors hindering the
scalability of WMI-based PFV, highlighting a number of promising
research directions.
Furthermore, through our unified perspective, we implement a
generalized version of bound propagation~\cite{dfv-nn-survey}, a
staple technique in neural network verification, that works for any ML
model supported by our formalism. We additionally implement an
iterative procedure that can potentially reduce the computational cost
of the verification by partitioning the search space into disjoint
problems.
We evaluated a prototypical WMI-based verifier implementing these two
techniques, obtaining a significant reduction in the verification
runtime.

The rest of the paper is structured as
follows. Sections~\ref{sec:related} and \ref{sec:background} discuss
the related work and introduce the required background respectively.
Our framework is presented in Section~\ref{sec:formulation}, followed
by a number of practical use cases where the PFV task reduces to WMI.
In Section~\ref{sec:motivating}, we showcase the importance of
verifying multiple properties of a model, even for very simple
systems.
We then investigate the challenges related to the scalability of the
approach in Section~\ref{sec:scalability} and introduce our prototypical
WMI-based verifier.
Section~\ref{sec:experiments} provides an empirical evaluation of the verifier on 
 a range of PFV tasks on neural networks and random forests.
We conclude by sketching a number of promising research direction
related to our work in Section~\ref{sec:conclusion}.

\section{Related Work}
\label{sec:related}
The common thread among the related work in ML verification is
targeting specific subclasses of problems (models and/or properties)
in order to achieve higher scalability.
For instance, model-specific algorithms have been devised for
tree-based models~\cite{dfv-trees},
continuous~\cite{reluplex,marabou} and binarized~\cite{dfv-bnn}
neural networks or support vector machines~\cite{dfv-svm}.
The largest body of work focuses on deterministic verification,
reducing the verification task to
decision~\cite{dfv-nn-safety,reluplex,marabou,dfv-trees2,dfv-gboost-rob},
or optimization~\cite{dfv-nn-rob-lp,dfv-nn-milp,dfv-trees-milp}
problems. This line of work cannot address quantitative tasks, like
verifying fairness properties or certifying nondeterministic systems.

Probabilistic model
checking~\cite{vardi1985automatic,katoen2016probabilistic}
traditionally addresses the verification of sequential systems,
achieving high scalability with strong distributional assumptions on
the priors~\cite{hartmanns2015quantitative}, such as independence or
unimodality.  For instance, random variables such as transition delays
in timed automata are usually modeled with independent exponential
distributions.
Therefore, these approaches are not suited for verification tasks
characterized by complex priors and highly combinatorial (ML) systems.
While our work focuses on non-sequential scenarios, we deem the
verification of sequential systems with ML components and complex
priors an important research goal.

The probabilistic verification of ML models is in its infancy. Most
works focus on verifying if specific properties, such as
robustness~\cite{pfv-nn-rob,pfv-nn-rob-proven,pfv-gnn-rob} or
fairness~\cite{pfv-fair-verifair}.
Closest to our framework is prior work on reducing quantitative
verification to combinatorial counting problems. This approach was
first applied to the evaluation of simple probabilistic programs with
univariate priors on random
variables~\cite{chistikov2015approximate,gehr2016psi}, and later
extended to fairness verification of decision procedures including
small ML models~\cite{pfv-fair-fairsquare}.  Similarly to our
approach, a probabilistic model of the environment is part of the
specification that the learned model must satisfy, albeit the only
property considered in this work is group fairness.  Quantitative
robustness verification of binarized NNs was also reduced to a
counting problem, leveraging the advances in approximate counting over
propositional logic theories~\cite{baluta2019quantitative}. This
approach is however limited to discrete systems. The framework
presented here is a generalization of the formalism above to
logical/numerical problems.

\section{Background on Weighted Model Integration}
\label{sec:background}

\begin{figure*}
  \centering
  \includegraphics[height=0.17\textwidth]{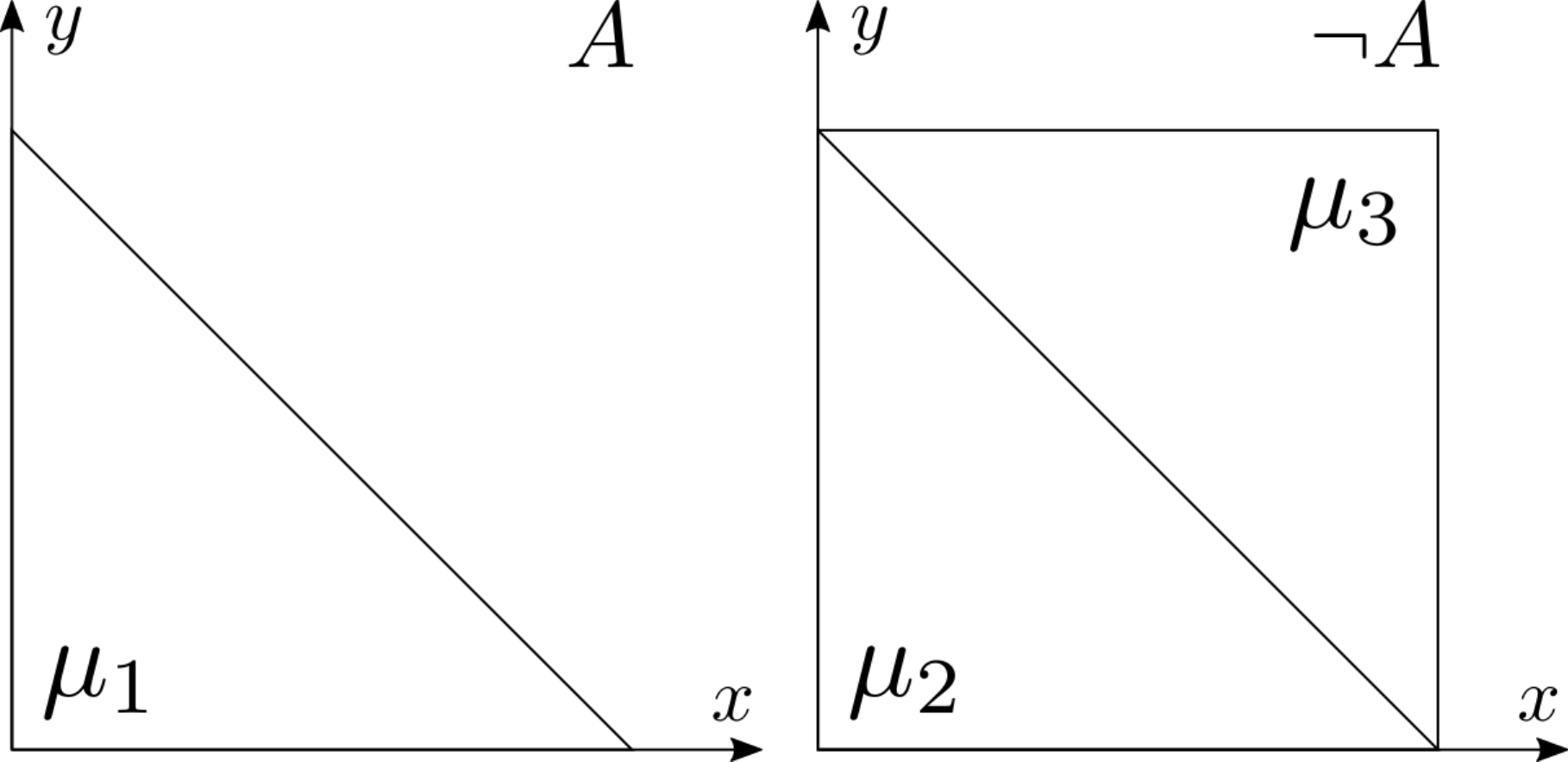}
  \hspace{0.3cm}
  \includegraphics[height=0.17\textwidth]{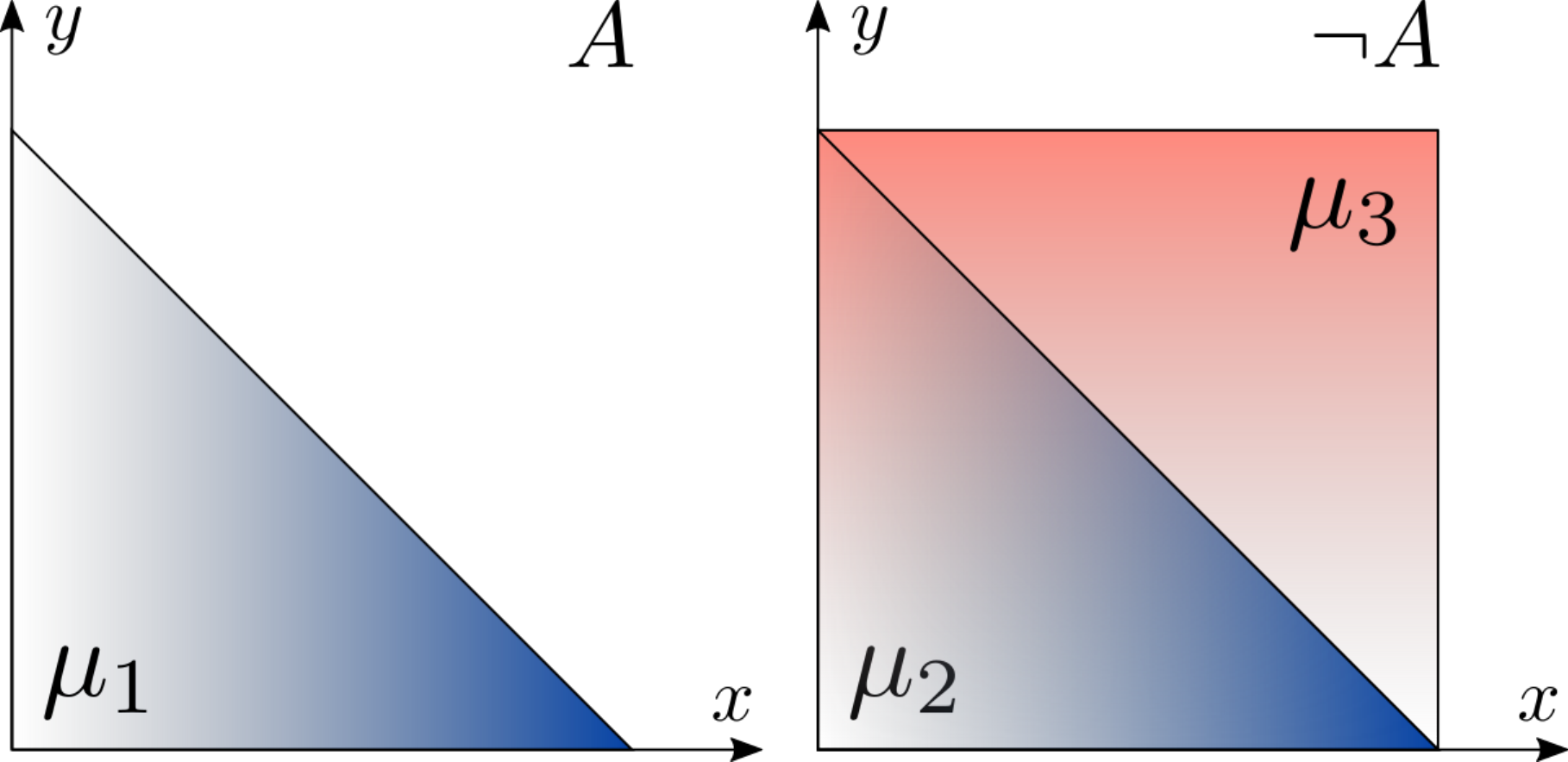}
  \hspace{0.3cm}
  \scalebox{0.53}{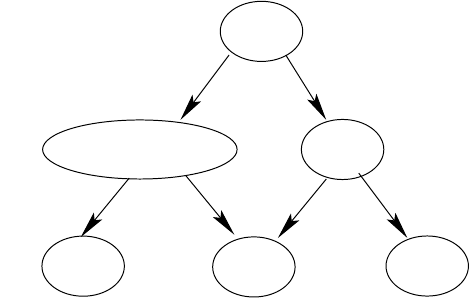}
  \caption{\label{fig:example} (Left) The hybrid region defined in
    Ex.~1.  (Center) The weighted formula in Ex.~2. (Right) An
    equivalent XADD representation of the weight.}
\end{figure*}

We assume the reader is familiar with the basic syntax, semantics, and
results of propositional and first-order logic. We adopt the notation
and definitions in \cite{morettin2019advanced} --- including some
terminology and concepts from Satisfiability Modulo Theories (SMT) ---
which we summarize below.  We will restrict SMT to quantifier-free
formulas over linear real arithmetic (\smtlra{}) which consist in
atomic propositions over some set \allA{} (aka {\em Boolean atoms}),
linear (in)equalities $(\sum_i a_ix_i\bowtie b)$ over a set of
continuous variables \allx{} s.t. $a_i,b$ are rational constants and
$\bowtie\ \in \set{\le,\ge,>,<,=,\neq}$ (aka \lra{} {\em atoms}), all
of them combined with the standard Boolean operators
\set{\neg,\wedge,\vee,\rightarrow,\leftarrow,\leftrightarrow}. (Our
work easily extends to the integers~\cite{kolb2018efficient}.)  The
notions of literal, clause, partial and total truth assignment extend
directly to \smtlra{} atoms.
We represent truth assignments $\mu$ as conjunctions of literals,
s.t. positive and negative literals in $\mu$ mean that the atom is
assigned to true and false respectively. 

For brevity, we introduce {\em interval formulas} $(x \in [l, u])
\defas (l \le x) \land (x \le u)$ and {\em if-then-else terms}
$\ite(.;.;.)$ s.t. $t=\ite(cond; e_1; e_2) \defas (cond \rightarrow
t=e_1) \land (\neg cond \rightarrow t=e_2)$.

\smtlra{} is concerned with finding a truth assignment $\mu$ to the atoms of
a formula $\Delta$ that both tautologically entails $\Delta$
($\mu\pmodels\Delta$) and is \lra-satisfiable.
We denote with
$\TTA(\Delta)$ the set of all the \lra{}-satisfiable total
truth assignments satisfying $\Delta$.
For instance, if $\Delta \defas (x < 0) \lor (x > 1)$, then
$\mu\defas (x < 0)\wedge(x > 1)\pmodels\Delta$ but it is not
\lra-satisfiable.
Thus, the formula has only two \lra-satisfiable total truth
assignments: $\TTA(\Delta) = \{(x < 0) \land \neg(x > 1), \neg(x < 0)
\land (x > 1) \}$.  (Hereafter, we may drop ``\lra-satisfiable'' when
referring to assignments in $\TTA(\Delta)$ since this is clear from context.)
Notice that every $\mu \in \TTA$ is a convex subregion of
$\Delta$.
We denote with $\mulra{}$ the portion on $\mu$ mapping \lra-atoms to truth values.
%
%
\begin{example}
The 
formula 
%
 $ \Delta \defas (x \in [0,1]) \land (y \in [0,1]) \land (A \rightarrow (x +
  y \le 1))$
%
defines a region over $\allA \defas \{A\}$ and $\allx \defas \{x,
y\}$, depicted in Figure~\ref{fig:example} (left). It results in 3
(total) \lra-satisfiable
truth assignments
(we omit $(x \in [0,1])\wedge(y \in [0,1])$ for short):
\begin{align*}
  \TTA(\Delta) = &\{ \overbrace{ A \land (x + y \le 1)}^{\mu_1},\\&
  \overbrace{\neg A \land (x + y \le 1)}^{\mu_2},
  \overbrace{\neg A \land \neg (x + y \le 1)}^{\mu_3} \}.
         \quad \diamond
\end{align*}
\end{example}
%


We introduce non-negative weight functions
$\w:\allx\cup\allA \mapsto \mathbb{R}^+$,
which intuitively defines a (possibly unnormalized) density function over $\allA \cup \allx$.
\wxa is defined by: (1) a \smtlra{} formula $\support$, called the \emph{support} of $w$, outside of which $w$ is $0$; (2) a combination of real functions whose integral over $\mulra$ is computable, structured by means of nested if-then-elses with \lra-conditions
%
(we refer to \cite{morettin2019advanced} for a formal definition of \wxa). 
When (2) is a DAG with polynomial leaves, as depicted in Fig.~\ref{fig:example} (right), it is equivalent to the notion of \emph{extended  algebraic decision diagram} (XADD) introduced
by
Sanner et al.~(2012).
%
%
$\wmu$ denotes \w{} restricted to 
the truth values of $\mu$.  
%
%
The {\bf Weighted Model Integral} of
a formula $\Delta(\allx,\allA)$ and a weight function
\w{}(\allx,\allA)  over $\support$ is defined as:
\begin{eqnarray}
  \label{eq:wmidef}
  \textstyle
 WMI(\Delta, \w) \defas   \sum_{\mu\in\TTA(\Delta\wedge\support)} \int_{\mulra} \wmuagen{\mu}(\allx) \: d\allx
\end{eqnarray}
where $\int_\Delta$ denotes integration over the solutions of a
formula $\Delta$.
WMI generalizes Weighted Model
Counting (WMC)~\cite{sang2005performing}.
As with WMC, WMI can be interpreted as
the total unnormalized probability mass of the pair $\langle \Delta, \w
\rangle$, i.e. its \emph{partition function}. The normalized
conditional probability of $\Gamma$ given $\Delta$ can then be
computed as:
\begin{align}
  \label{eq:normwmi}
  \cprob{\Gamma}{\Delta} = WMI(\Gamma \land \Delta, w) / WMI(\Delta, \w).
\end{align}

\begin{example}
  Let $\Delta$ be as in Ex.1,
  with
  weights:
\begin{align*}
  &w_{(x + y \le 1)} (x, y) = x, \quad\quad w_{\neg A} (x, y) = y, \\&w_\ell(x,y) = 1 \quad\quad \forall \ell \notin \{(x+y \le 1), \neg A\}
\end{align*}
That is,  $w$ defines the following piecewise density over
the convex subregions of $\Delta$ (Fig.~\ref{fig:example}, center):
\begin{align*}
  \wxa\defas
  ite((x+y\le 1);x;1)\times
  ite(A;1;y)
\end{align*}
with $\support\defas\top$, so that:
$\left \{\begin{array}{ll} 
  \wmuagen{\mu_1}(x, y) &= 
  x \times 1 = x\\
  \wmuagen{\mu_2}(x, y) &= 
  x \times y= xy\\
  \wmuagen{\mu_3}(x, y) &= 
  1 \times y = y
\end{array}\right \}$.\\

Then, the weighted model integral of
$\langle \Delta, w
\rangle$ is:
\begin{align*}
  WMI(\Delta, w) &= \int_0^1\!\!\!\int_0^{(1-y)}\!\!\!x \:dx \:dy + 
\int_0^1\!\!\!\int_0^{(1-y)}\!\!\!xy \:dx \:dy \\&+ \int_0^1\!\!\!\int_{(1-y)}^1\!\!\!y \:dx \:dy
= \frac{1}{6} + \frac{1}{24} + \frac{1}{3} 
= \frac{13}{24}.
\end{align*}
We can now compute normalized probabilities such as $$\cprob{A}{\Delta} =
\frac{WMI(A \land \Delta, w)}{WMI(\Delta, w)} = \frac{1/6}{13/24} =
\frac{4}{13} \quad \diamond$$ 
\end{example}

Crucially for our purposes, $\Gamma$ and $\Delta$ in
\eqref{eq:normwmi} can be \emph{arbitrarily complex} SMT-LRA
formulas. This gives as a very powerful computational tool for
quantifying the probability associated to complex events or
properties. 

\section{WMI-based ML verification}
\label{sec:formulation}
In this section, we first introduce our unifying perspective on the
PFV of ML systems, trying to model the problem in the most general
terms.
Then, we show how the computational task that we define can be reduced
to WMI for a broad range of use cases, addressing the desiderata
introduced in Section~\ref{sec:intro}.

\subsection{Our unified theoretical perspective}

Without loss of generality, we denote the system under verification
with $\bY \sim \psys(\bY | \bX)$, mapping inputs $\bX$ into outputs
$\bY$ via an arbitrary conditional distribution, with generic
deterministic systems $\bY = f(\bX)$ being special cases of our
formulation.
The requirement $\pro$ is composed of a pair of formal statements: a
\emph{precondition} $\propre$, typically defined over $\bX$ only, and
a \emph{postcondition} $\propost$, which is typically defined over
$\bY$.
If a property of the output is expected to hold on the whole input
domain, we set $\propre \defas \top$.
In contrast with deterministic verification, $\pro$ additionally
includes a \emph{prior distribution} $\ppro(\bX)$ over the input
variables, probabilistically modeling the environment where the
system is going to be deployed.
If no informed prior is available in the specification, $\ppro$ is the
uniform distribution $\mathcal{U}(.)$.

Then, given a probability threshold $k \in [0, 1]$, we are concerned
with verifying that $\propost$ holds with at least probability $k$
whenever $\propre$ holds:
\begin{align}
  \label{eq:pfvtask}
  &\cprob{\propost}{\propre, \sys} \geq k \\
  \text{with }\quad &\bX,\bY \sim \prob{\bX, \bY} \defas \ppro(\bX) \cdot \psys(\bY \:|\: \bX) . \nonumber
\end{align}
Besides naturally supporting the verification of probabilistic
systems, we argue that the task in \eqref{eq:pfvtask} has a practical
application even when $\sys$ is deterministic.
For instance, properties like \emph{demographic parity} and
\emph{individual fairness} are both based on the notion of a
population model, i.e. a distribution over individuals. In our
framework, this is modeled by $\ppro$.
Moreover, while many properties do not \emph{always} hold in practical
scenarios, it is perfectly acceptable for them to hold with high
probability (or low, for undesired properties). This might be the case
when we are concerned with the robustness of a classifier to random
noise, rather than a rational attacker. In Section~\ref{sec:exp-rob},
we consider a probabilistic notion of local robustness, discriminating
highly non-robust instances from the ones characterized by an
acceptable probability of changing prediction (i.e. $< k$) in their
neighborhood.

We remark that, while \emph{deterministic} formal verification (DFV)
can be retrieved by setting $k = 1$, in general solving the PFV task
in \eqref{eq:pfvtask} requires quantitative reasoning capabilities
that make the two problems fundamentally different.
%
%
We rewrite the LHS term of \eqref{eq:pfvtask} as follows, to make the
role of $\ppro, \psys, \propre$ and $\propost$ explicit:
\begin{align*}
  \cprob{\propost}{\propre, \sys} &= \frac{\cprob{\propost, \propre}{\sys}}{\cprob{\propre}{\sys}} \\ 
  &= \frac{\int \ind{\propre \land \propost}\quad \psys(\bY|\bX) \cdot \ppro(\bX) \: d\bX \: d\bY}
        {\int \ind{\propre} \quad \psys(\bY|\bX) \cdot \ppro(\bX) \: d\bX \: d\bY} \nonumber
\end{align*}
where $\ind{\cdot}$ denotes the indicator function. From
\eqref{eq:normwmi}, it follows that the above can be computed via
WMI:
\begin{align}
  \label{eq:pfvwmi}
 \cprob{\propost}{\propre, \sys} &= \frac{WMI(\encpre \land \encpost, w)} {WMI(\encpre, w)}
\end{align}
as long as $\psys \cdot \ppro$ can be modeled as a structured weight
function $w = w_\sys \cdot w_\pro$ belonging to the family introduced
in Section~\ref{sec:background}, and $\propre$ and $\propost$
admit \smtlra encoding, indicated as $\encpre$ and $\encpost$ respectively.

We show that this is the case for a many probabilistic
(Sec.~\ref{sec:pmodelling}) and deterministic
(Sec.~\ref{sec:mlmodelling}) ML models, as well as many properties
(Sec.~\ref{sec:properties}) of practical interest.
Since SMT encodings can be naturally conjoined and the class of weight
functions supported by the formalism is closed under
multiplication~\cite{wmisurvey}, it is trivial to combine the
encodings of $\sys$ and $\pro$ from both logical and probabilistic
perspectives.

\subsection{Modeling arbitrary distributions (\textbf{D1})}
\label{sec:pmodelling}

\customparagraph{Simple densities.}  In WMI, the choice of density
functions is dictated by the integration procedure adopted by the
solver. In general, any function can be used as long as it is
integrable,
either 
exactly or approximately, in a region defined by a
conjunction of \lra atoms, i.e. a convex polytope.
Most exact WMI algorithms support piecewise polynomials as building
blocks for the structured weight functions. The reason is twofold: 1)
polynomials are easy to work with, being closed under sum, product and
integration over polytopes; 2) they can approximate any density with
arbitrary precision.
Notably, the complexity of exact integration grows only polynomially
with respect to the degree~\cite{baldoni2011integrate}, making
higher-degree piecewise functions sometimes preferable over a
drastically higher number of lower-degree pieces.
When Monte Carlo-based procedures are used, any non-negative density
can be adopted, including Gaussians~\cite{merrell2017weighted,
  dos2019exact}, as long as statistical guarantees on the
approximation are provided. \\

\begin{figure*}
  \centering
  \includegraphics[height=0.18\textwidth]{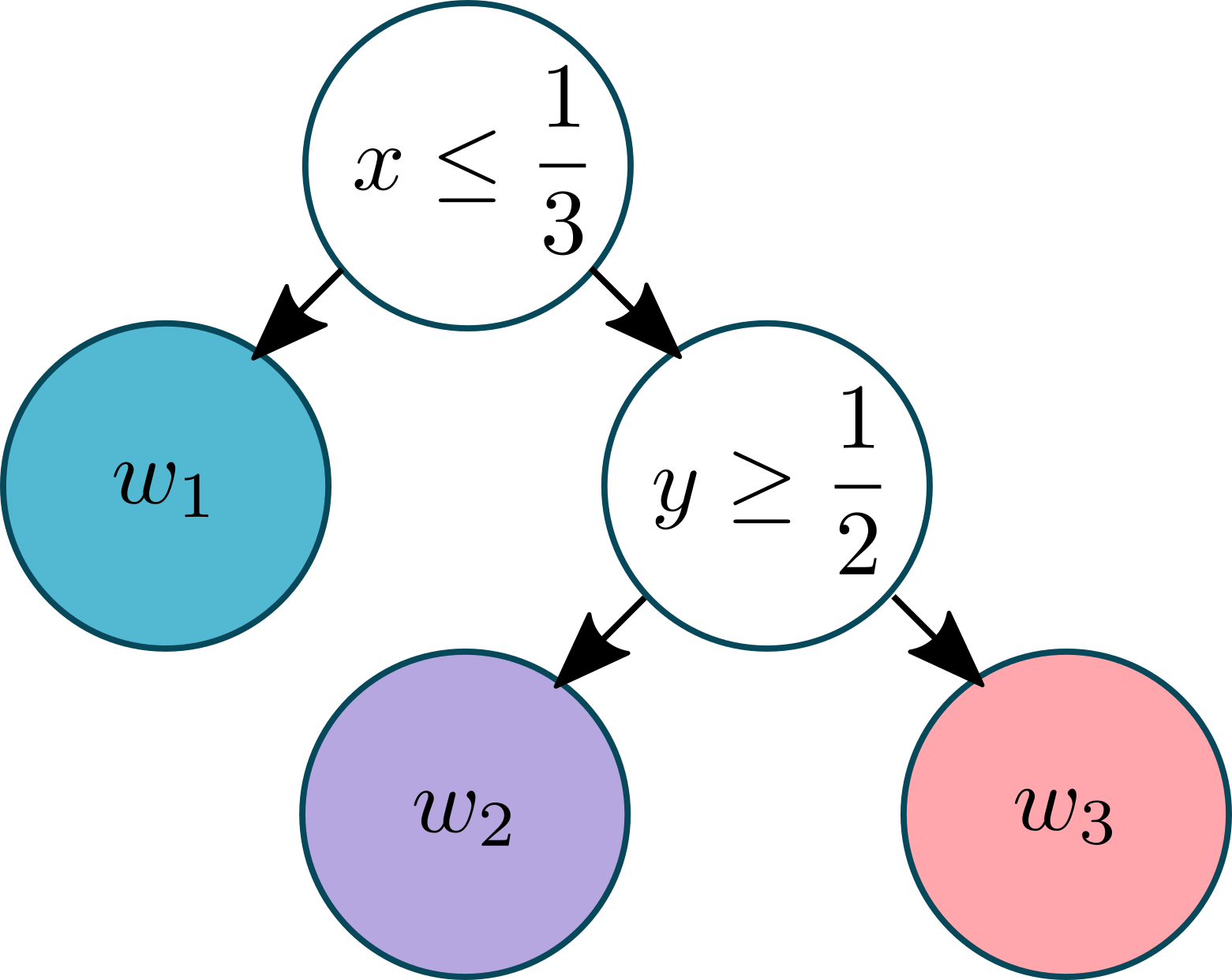}
  \includegraphics[height=0.18\textwidth]{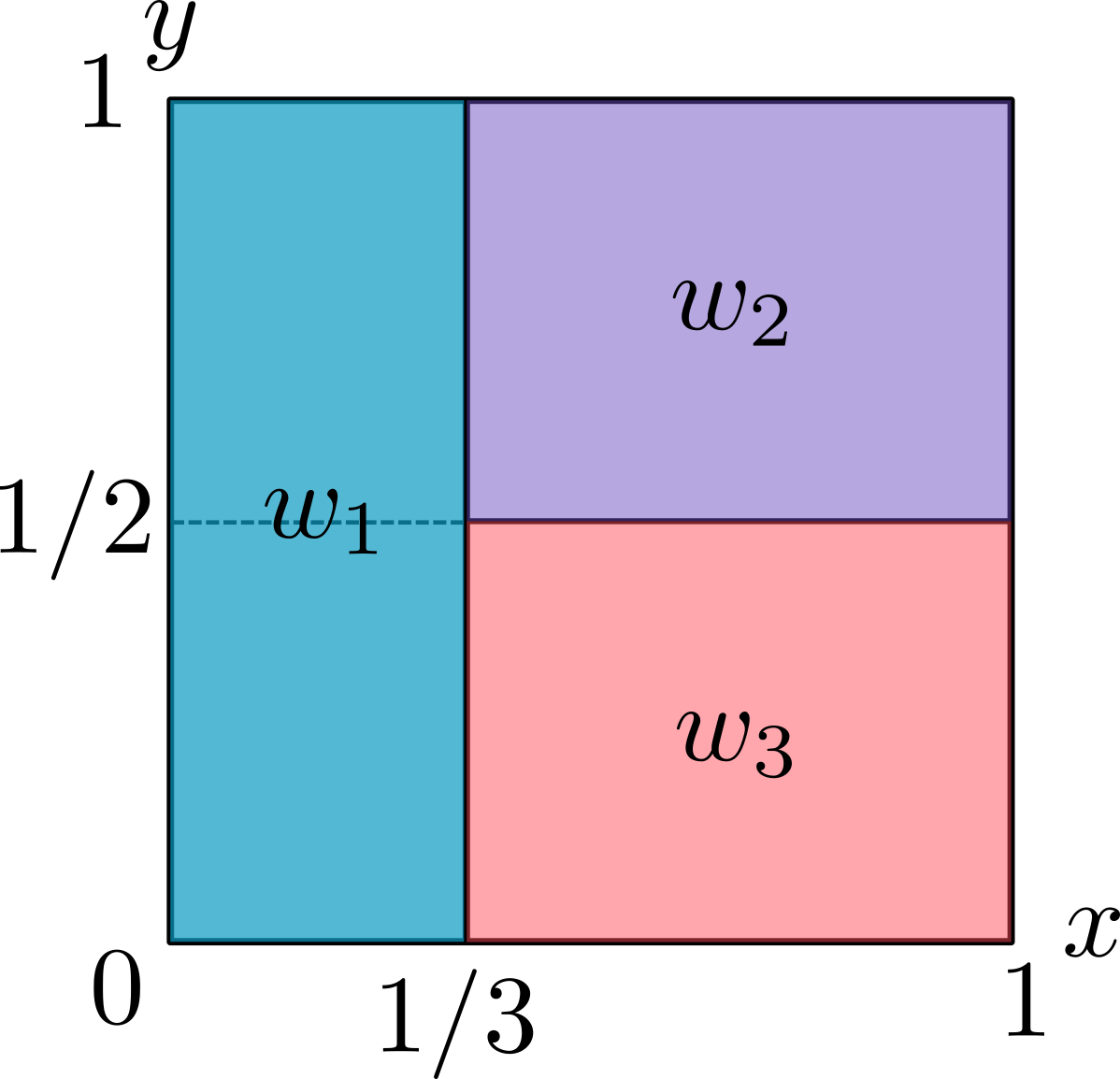}
  \hspace{1.0cm}
  \includegraphics[height=0.18\textwidth]{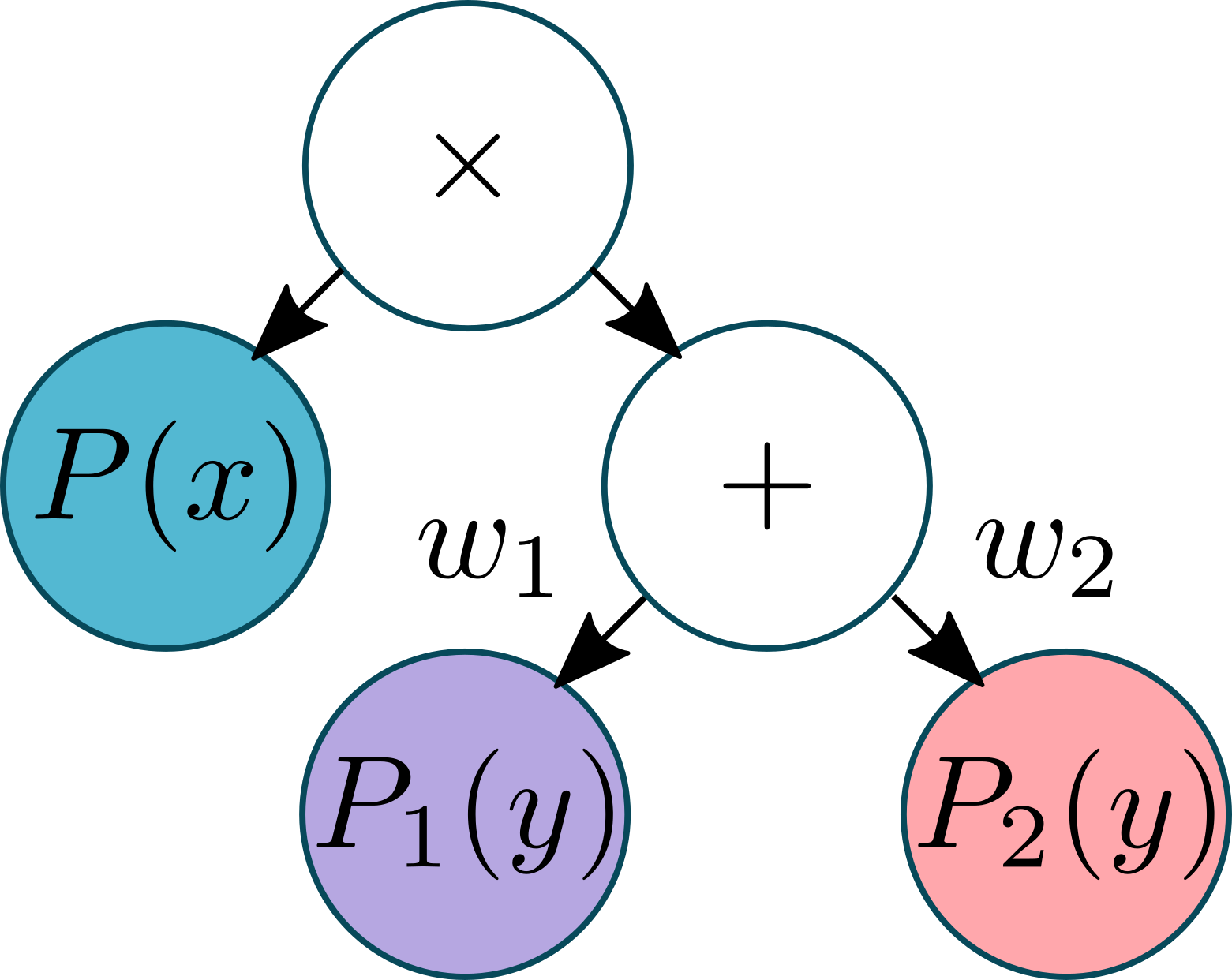}
  \includegraphics[height=0.18\textwidth]{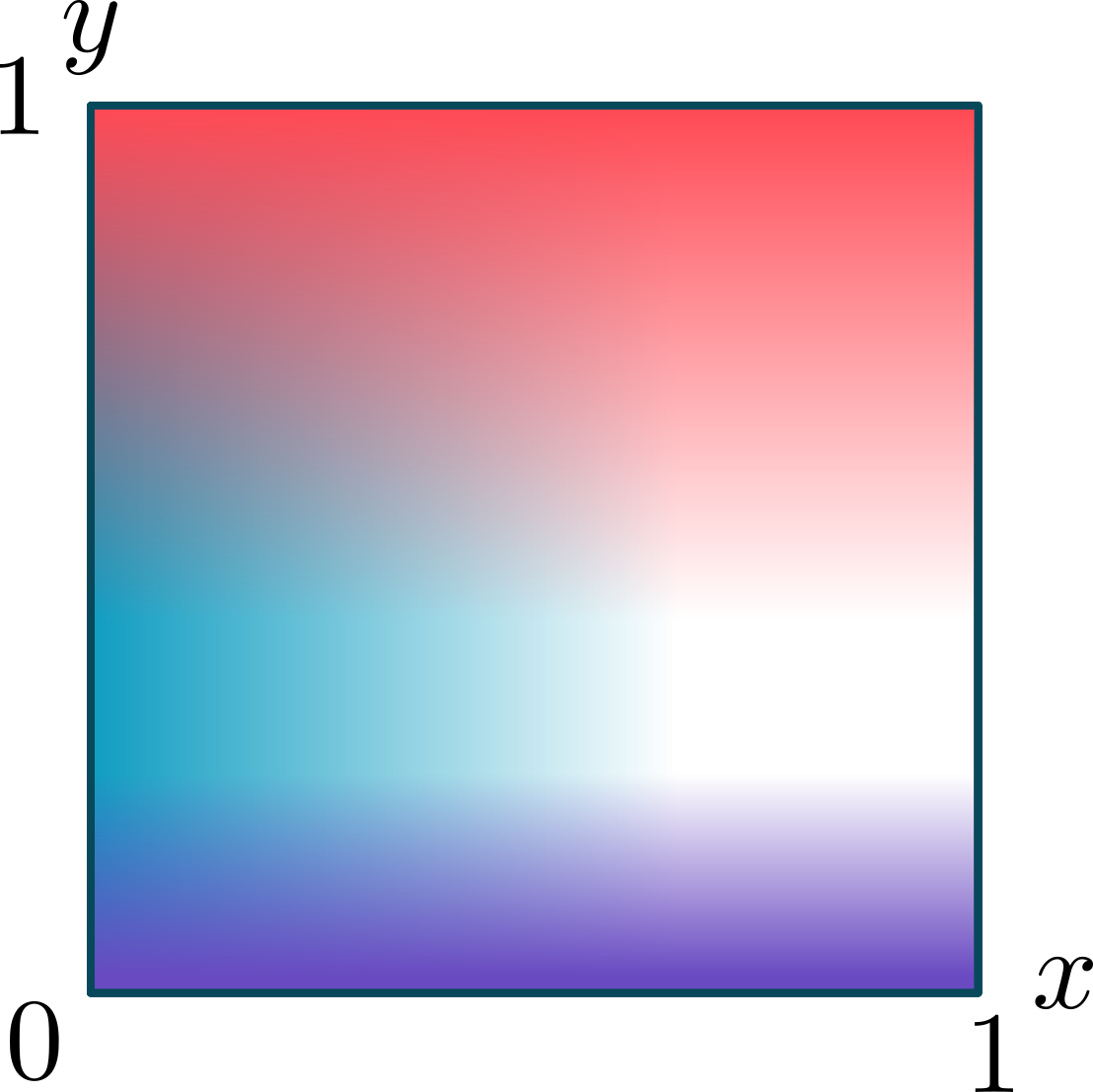}
    \caption{\label{fig:detspn} Two structured densities supported by
      our framework: DETs (\emph{left}) and SPNs (\emph{right}).}
\end{figure*}

\customparagraph{Structured densities.}
By combining these density functions by means of conditional
statements, sums and products, the structure of a variety of popular
probabilistic models can be encoded. For illustrative purposes, we
report the encodings of two opposite approaches (depicted in
Fig.~\ref{fig:detspn}), one exclusively using conditional statements
in the structure and one not using them at all.
%
The former are \emph{Density estimation trees
(DETs)}~\cite{ram2011density}, the density estimation variant of
decision or regression trees. The hierarchical structure is composed
of binary decision nodes having univariate conditions $(x_i \le k)$ or
logical propositions $A_i$. Different from its classification and
regression variants, each leaf of a DET encodes the probability mass
corresponding to the subset of the support induced by the decisions in
its path from the root. The structure of DETs is therefore encoded as
a tree of $\ite$ with constant terms at the leaves.
The latter are \emph{Sum-product networks (SPNs)}~\cite{poon2011sum},
probabilistic circuits combining tractable univariate distributions by
means of mixtures (weighted sums) over the same variables or
factorizations (products) over disjoint sets of variables, resulting
in a tractable but expressive joint probability.
The common choices for the continuous distributions include Gaussians
and piecewise polynomials~\cite{molina2018mixed}, both amenable to
WMI inference.

Related work in WMI covers a number of
  probabilistic models falling in between these two opposite
  approaches, including (hybrid variants of) graphical models like
Bayesian~\cite{pfv-fair-fairsquare} and
Markov~\cite{belle2015probabilistic} networks, and probabilistic
logic programs with continuous variables \cite{dos2019exact}.

Typically, the support of these models is an axis-aligned bounding
box, either defined by the user or inferred from the training
data. Nonetheless, prior work in the WMI literature has been concerned
with integrating probabilistic ML models with complex and possibly
non-convex constraints~\cite{morettin2020learning}.

\subsection{Modeling deterministic ML systems (\textbf{D2})}
\label{sec:mlmodelling}

Deterministic systems are a use case of particular interest in ML
verification.
In our framework, these can be modeled by setting the weight term
$w_\sys = 1$ and fully encoding their behavior in its support
$\support_\sys$.
A broad range of ML models can be encoded in this way. Some
representative examples are reported below.\\

\customparagraph{Tree-based models.}
As shown above for DETs, tree structures can be encoded by means of
nested $\ite$ with propositional or linear conditions, with the
only difference that children of conditional expressions are formulas
rather than terms.
Then, leaves are simply encoded with (a conjunction of) atoms $(y =
c_i)$, mapping output variables to their respective values.
If axis-aligned splits are not expressive enough, arbitrary
\lra-conditions enable more complex piecewise decompositions of the
joint density, such as those employed in Optimal Classification
Trees~\cite{bertsimas2017optimal}. \\

\customparagraph{Neural Networks and Support Vector Machines.}
Piecewise-linear models can be exactly encoded via
  $\ite$.
For instance, ReLU NNs can be modeled by conjoining the encoding of
each unit $i$ (with inputs $\bX_i$, weights $\bW_i$ and $b_i$):
\begin{align*}
    (h_i = (\bW_i \bX_i + b_i)) \land
    (y_i = ite(h_i > 0, \: h_i, \: 0)) .
\end{align*}
Common operations like convolutions and max/average pooling have
\smtlra{} representations.
For classifiers, if the properties under inspection do not involve the
confidence of the predictor, encoding the last sigmoid can be avoided
by defining the property over the logits.
Other non-linear predictors, such as support vector machines with
piecewise linear feature maps \cite{pwsvm} can be
similarly encoded. Other classes of
  non-linearities, such as sigmoids, can be arbitrarily approximated~\cite{global2safetynn}. \\

\customparagraph{Ensemble models.}
The compositional nature of SMT 
can support ensembles of $K$ ML models,%
\begin{eqnarray}
  \nonumber \textstyle
   (\bY = a(\bY_1, ..., \bY_K)) \land \bigwedge_{i=1}^K \support_{f_i}(\bX) ,
\end{eqnarray}
as long as the aggregation function $a$ can be encoded. This is the
case for common choices like averaging or majority voting.

Two important considerations are in order. First, not being restricted
to a specific model family, this verification framework can include
any component that is encodable in \smtlra{}.
%
This is in stark contrast with many approaches in the literature, which
can only verify ML components in isolation.
 \\Second, by combining \smtlra formulas with structured
  weight functions as a unified representation formalism, our
  framework enables the use of \emph{certified} ML models as part of
  the specification, a fundamental feature when defining it manually
  is hard or impossible, like in image domains.
For instance, our framework encompasses a probabilistic version of
\emph{neuro-symbolic verification}~\cite{xie2022neuro}, where
certified NNs are used as high-level predicates in the property
itself, such as \emph{``A stop sign is in proximity''}.
We leave the investigation of ML-based specifications for future work,
focusing on more standard properties in the ML verification
literature.

\subsection{Modeling $\propre$ and $\propost$ (\textbf{D3})} 
\label{sec:properties}

In the following, we showcase the flexibility of the framework in
terms of properties that can be verified.
For classifiers, we denote with $c(\bX)$ the boolean variable in
$\support_\sys$ encoding a positive classification for $\bX$. If
$\support_\sys$ encodes a regressor, we denote with $f(\bX)$ the term
representing its continuous output.
Distances among input or output points are key for defining many
properties of interest.  Fortunately, $L_1$ and $L_\infty$ admit an
\smtlra encoding by means of nested if-then-elses:
\begin{align*}
\| \bX - \bX' \|_1 &\defas \sum_{i=1}^N \: |x_i - x_i'| \\ \| \bX -
\bX' \|_\infty &\defas \maxsmt_{i=1}^N \: |x_i - x_i'|
\end{align*}
where
\begin{align*}
  |v_1 - v_2| &\defas \ite(v_1 < v_2 ; \: v_2 - v_1 ;\: v_1 - v_2)\\
  \maxsmt(\{v_1, v_2\}) &\defas \ite(v_1 < v_2;\: v_2;\: v_1)\\
  \maxsmt(\{v_1, v_2\} \cup V) &\defas \ite(v_1 < v_2;\: \maxsmt(\{v_2\} \cup V);\\
  &\quad\quad\quad\quad\quad\quad\quad \: \maxsmt(\{v_1\} \cup V)) .
\end{align*}

\customparagraph{Dimensionality-preserving properties.}
The properties falling into this category preserve the
dimensionality of the integration. As discussed in
  Section~\ref{sec:scalability}, this is a crucial factor affecting
  the complexity of the verification task.
From the above, it follows that we can encode \emph{local
classification robustness} around the input point $\bX_0$ (having
class $c_0$) as:
\begin{align}
  \label{eq:localrob}
  \encpre &\defas \|\bX - \bX_0 \| \le \epsilon \\
  \encpost &\defas (c(\bX) \leftrightarrow c_0) \nonumber\\
  w(\bX, \bY) &\defas w_\pro(\bX) \cdot w_\sys(\bX, \bY). \nonumber
\end{align}
Robustness (and any other property that involves checking the
similarity among outputs) of regressors can be similarly encoded by
considering a distance among predictions:
$$\encpost \defas (\|f(\bX) - y_0 \| \le \delta) .$$

Other distance-based global properties include verifying the
probabilistic \emph{equivalence of two predictors} $\sys_1$ and
$\sys_2$:
\begin{align}
  \label{eq:equivalence}
  \encpre &\defas \top \\
  \encpost &\defas (c_{\sys_1}(\bX) \leftrightarrow c_{\sys_2}(\bX)) \nonumber\\
  w(\bX, \bY) &\defas w_\pro(\bX) \cdot w_{\sys_1}(\bX, \bY) \cdot w_{\sys_2}(\bX, \bY) . \nonumber
\end{align}
Crucially, $\sys_1$ and $\sys_2$ can be belong to different families
of ML models, e.g. a neural network and an ensemble of DTs.

Group fairness notions like \emph{demographic parity}:
\begin{align}
  \label{eq:parity}
  r_{parity} = \frac{\cprob{c(\bX)}{\bX \in \mathcal{M}}}{\cprob{c(\bX)}{\bX \not\in \mathcal{M}}}
\end{align}
require computing \emph{ratios} of conditional probabilities, which
can be accommodated with a slight modification
of~\eqref{eq:pfvtask}. Our framework support these use cases as long
as the membership to a protected minority $\Delta_{\mathcal{M}} \defas
\bX \in \mathcal{M}$ can be encoded:
\begin{align}
  \label{eq:gfair}
  \encpre &\defas \Delta_{\mathcal{M}} \\
  \encpost &\defas c(\bX) \nonumber\\
  w(\bX, \bY) &\defas w_\pro(\bX) \cdot w_\sys(\bX, \bY) . \nonumber
\end{align}
Then,
$$r_{parity} = \frac{WMI(\encpre \land \encpost, w) / WMI(\encpre, w)}{WMI(\neg \encpre \land \encpost, w) / WMI(\neg \encpre, w)} .$$
This is the case when the set $\mathcal{M}$ is characterized by a
combination of categorical features, arguably the most common case in
fairness scenarios. Additionally, numerical constraints can be encoded
with arbitrary piecewise linear sets. \\

\customparagraph{Hyperproperties.}
Many properties of interest require evaluating the system multiple
times or, from a probabilistic viewpoint, multiple instantiations of
the same random variables, e.g. $\bX, \bX' \sim \prob{\bX}$.
These are known as hyperproperties in the program analysis
literature~\cite{barthe2011secure}, and are evaluated by means of
\emph{self-composition}, i.e. composing the encoding of the original
system with a copy where variables have been renamed.
This approach has recently been used for the verification of
properties of NNs both in deterministic~\cite{global2safetynn} and
probabilistic~\cite{pfv-nn-rob} scenarios.

\begin{figure}
  \centering
  \includegraphics[width=0.20\textwidth]{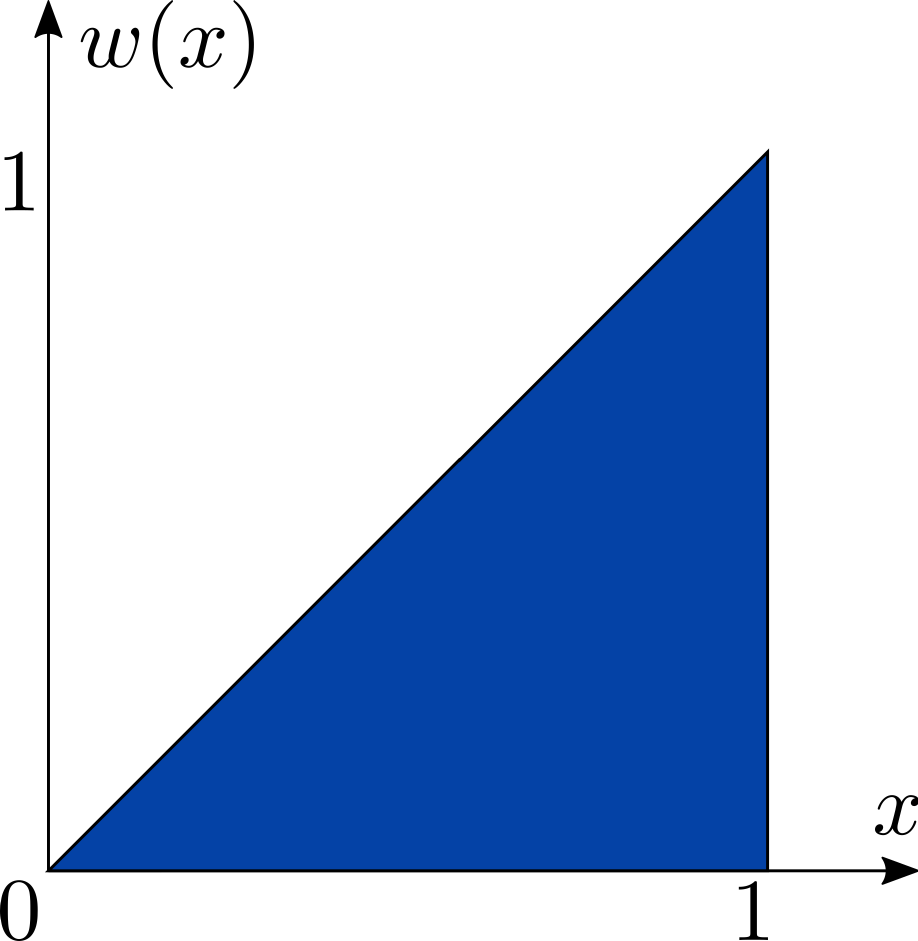}
  \hspace{0.3cm}
  \includegraphics[width=0.20\textwidth]{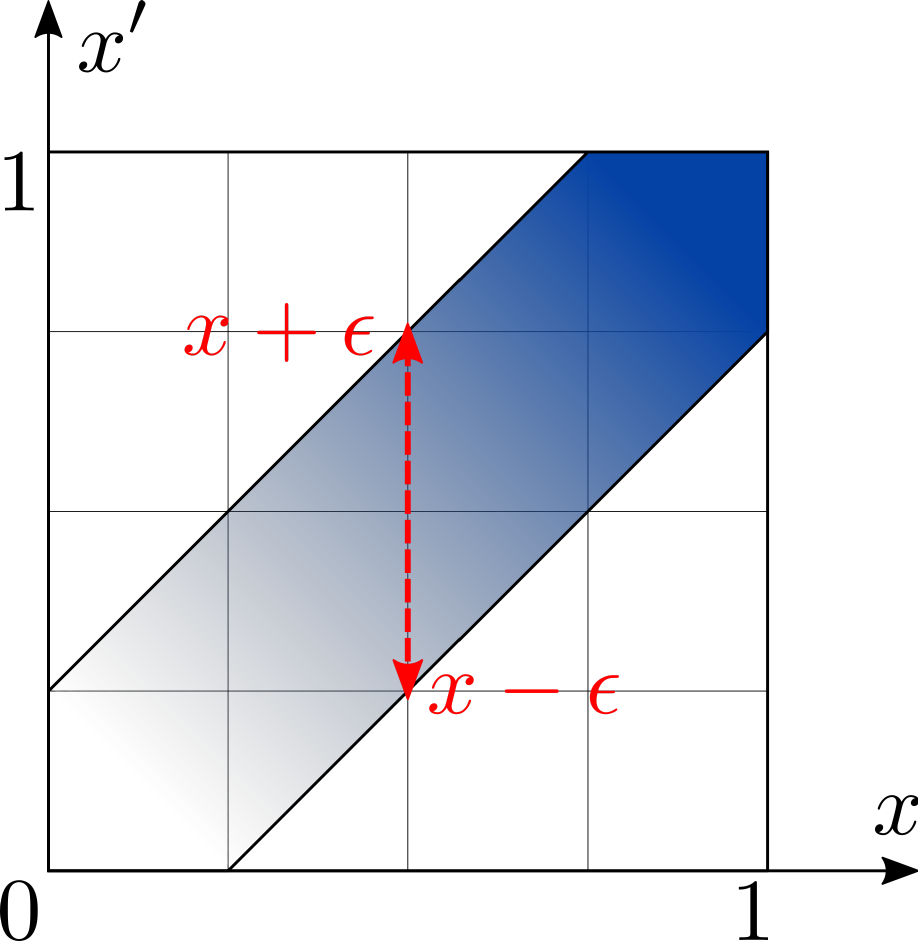}
  \caption{\label{fig:copy}  (\emph{Left}) A simple distribution $\Delta = (x \in [0,1])$ and $w(x) = x$.  (\emph{Right}) the distribution obtained by self-composition, enabling the computation of queries like
    $\prob{x' \in [x-\epsilon,x+\epsilon]}$.}
\end{figure}

Our framework generalizes the work above to arbitrary probabilistic ML
models, by simply defining self-composition over weight terms and
their support:
\begin{align*}
  \scomp(\chi) \defas \chi \land \chi[v \leftarrow v'], \quad\quad \scomp(w) \defas w \cdot w[v \leftarrow v'] ,
\end{align*}    
where $[v \leftarrow v']$ denotes the substitution of all variable
with fresh ones.
This approach guarantees that $\bX$ and $\bX'$ are independently drawn
from the same distribution.
We can then compute queries
involving both, such as $x' \in [x-\epsilon, x+\epsilon]$ 
(Fig.~\ref{fig:copy}).

For instance, our framework can leverage self-composition for verifying
\emph{individual fairness} of a classifier:
\begin{align*}
  \encpre &\defas \|\bX - \bX'\| < \epsilon \\
  \encpost &\defas (c(\bX) \leftrightarrow c(\bX')) \\
  w(\bX, \bY) &\defas \scomp(w_\pro(\bX) \cdot w_\sys(\bX, \bY)),
\end{align*}    
or the \emph{monotonicity} of a regressor:
\begin{align}
  \label{eq:mono}
  \encpre &\defas (\bX < \bX') \\
  \encpost &\defas (f(\bX) < f(\bX')) \nonumber\\
  w(\bX, \bY) &\defas \scomp(w_\pro(\bX) \cdot w_\sys(\bX, \bY)) . \nonumber
\end{align}
Notice that, in the two use cases above, self-composition is applied
to the encoding of both $\psys$ and $\ppro$.
Sometimes, it is enough to self-compose $\psys$, for instance when
quantifying how \emph{robust} a classifier is to some noise $\prob{\bN}$:
\begin{align}
  \label{eq:robnoise}
  \encpre &\defas (\bX' = \bX + \bN) \\
  \encpost &\defas (c(\bX) \leftrightarrow c(\bX')) \nonumber\\
  w(\bX, \bY) &\defas w_\pro(\bX) \cdot w_\pro(\bN) \cdot \scomp(w_\sys(\bX, \bY)) . \nonumber
\end{align}

\section{Use case: an income classifier}
\label{sec:motivating}
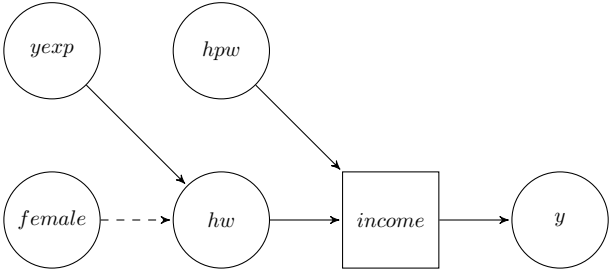
\begin{figure}[h]
  \centering
  \begin{tikzpicture}[>=stealth', shorten >=1pt, auto,
      node distance=1.2cm, scale=0.8, 
      transform shape, align=center, 
      observed/.style={circle, draw, minimum size=1.6cm},
      latent/.style={rectangle, draw, minimum size=1.6cm}]

    \node (yexp)  [observed] {$yexp$};
    \node (hpw)  [observed, right=of yexp] {$hpw$};
    \node (hw)  [observed,below=of hpw] {$hw$};
    \node (female) [observed, left=of hw] {$female$};
    \node (income)  [latent,right=of hw] {$income$};
    \node (y)  [observed,right=of income] {$y$};

    \draw[->,dashed] (female) to (hw);
    \draw[->] (hpw) to (income);
    \draw[->] (yexp) to (hw);
    \draw[->] (hw) to (income);
    \draw[->] (income) to (y);
  
  \end{tikzpicture}
  \caption{
    \label{fig:incomegraph}
    Graphical model of the generating distribution. The
    observed variables in the sampled data are circled. The dependency
    $female \rightarrow hw$ is only present in the biased dataset.}
\end{figure}

\emph{Is the ability of formally verifying multiple properties of ML
models really important?}
This is a legitimate question in particular for relatively simple
systems, whose behavior should be more easily predictable.
To answer it, we devised a prototypical scenario involving a seemingly
trivial classification task.
We then quantified multiple properties while training the models,
reflecting a different perspective on their
correctness.
Despite its apparent simplicity and the high empirical performance of
the systems, this motivating use case showcases the dangers of
narrowing the evaluation of a system to a few criteria.
Taking inspiration from the Adult UCI dataset, we devised a binary
classification task where $\sys$ has to predict whether an
individual's yearly income is greater than $35K$ dollars or not, based
on three features: the working hours per week ($hpw$), the years of
working experience ($yexp$) and the hourly wage ($hw$), as depicted in Figure~\ref{fig:incomegraph}.

\begin{figure}[h]
  \centering
  \hspace{-0.7cm}\includegraphics[width=0.5\textwidth]{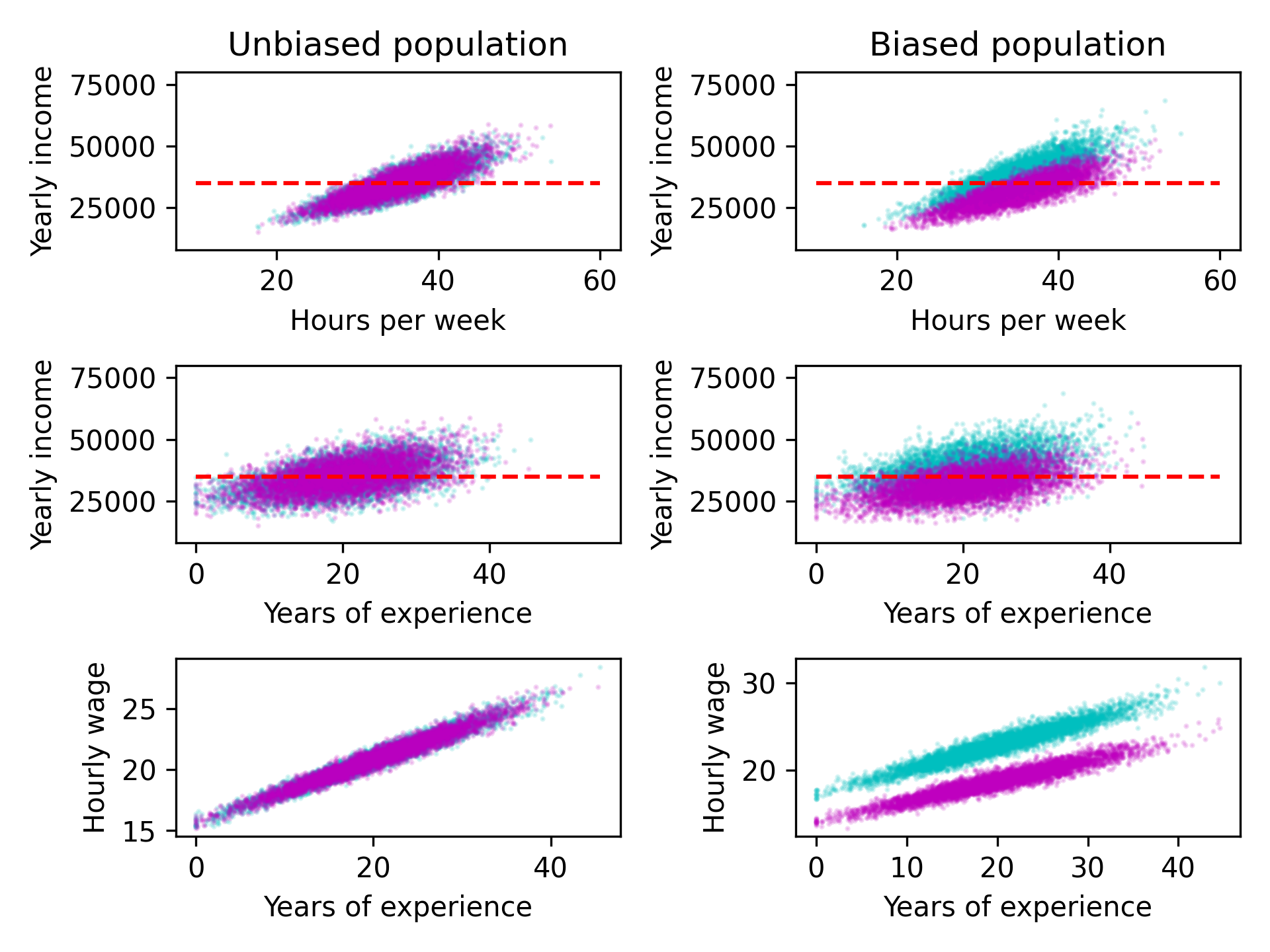}
  \caption{
    \label{fig:pop-data}
    Data distributions used to train the population model, plotted
    using different colors for $female$ and $\neg female$ instances.}
\end{figure}

Additionally, our ground-truth generative model includes a binary
variable $female$ and a toggleable mechanism that statistically
penalizes the hourly wage of female individuals
(Figure~\ref{fig:pop-data}).

We considered both biased and unbiased scenarios, each having separate
training datasets for the population model and the system under
verification, with $10000$ instances each (for details on the
distribution, see Section~\ref{sec:app-motivating}).

We trained we a DET as a population model, serving as a prior $\ppro(female,
hpw, hw, yexp)$.
We implemented $\sys$ as a very simple ReLU NN classifier $f(hpw, hw,
yexp)$ in PyTorch~\cite{pytorch2019} with two hidden layers of size
$8$, trained using stochastic gradient descent with the objective of
minimizing the binary cross-entropy loss.
Despite its size, $f$ was expressive enough to achieve high predictive
performance, reaching an accuracy of $0.9$ on both validation and test
set around epoch $300$ in both the biased and unbiased scenarios.

Throughout the training of $\sys$, we quantified multiple properties
by reducing the computation of the LHS in \eqref{eq:pfvtask} to WMI
and leveraging the off-the-shelf exact solver
SA-WMI-PA~\cite{spallitta2022smt}.
We monitored the demographic parity~\eqref{eq:gfair} and the
robustness of $f$ to noise in the input~\eqref{eq:robnoise}, the
latter property reflecting the assumption that individuals tend to
report inflated positive attributes.
For illustrative purposes, we modeled an additive noise to $yexp$
with a triangular distribution with mode $0$ and support $[0, 15]$ and
computed how likely the output would not change in spite of
statistically incorrect input.
Since the income is clearly monotonic with respect to the input
features, we additionally verified the correctness of $f$ by
quantifying its monotonicity~\eqref{eq:mono} with respect to each
feature. Given two inputs $\bX$ and $\bX'$ differing on a single
feature $i$, we measured how likely it is that $f(\bX) > f(\bX')$ if
$x_i > x_i'$. Since this mathematical property should hold regardless
of $\ppro$, we switched $\ppro$ for a uniform prior.

\begin{figure}
  \centering
  \includegraphics[width=0.45\textwidth]{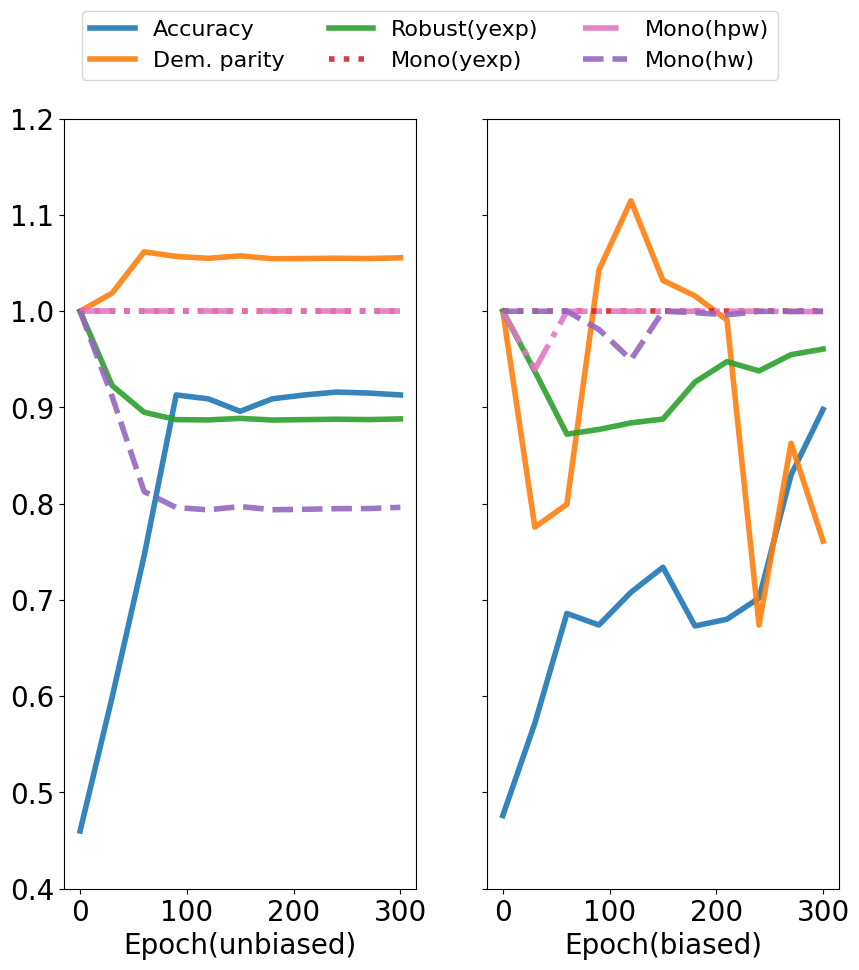}
    \caption{\label{fig:toy} 
    Quantitative evaluation while training in the unbiased
    (\emph{left}) and biased (\emph{right}) scenarios. Monotonicity
    results mostly overlap, indicating that the NN correctly learned
    the relation, except for \emph{hours per week} in the unbiased
    scenario.}
\end{figure}

Despite the simplicity of the predictive task, our results clearly
show that looking at the empirical accuracy alone or at a single
formal property might deceive the user into considering $f$ to be
correct~(Fig.~\ref{fig:toy}).
The monotonicity with respect to $hw$ of the model trained on unbiased
data is quite low. Since we used a uniform prior in the computation,
this implies that in 20\% of input domain volume the monotonicity
constraint is not respected.
Moreover, the network is not as robust to noise in $yexp$ as the
biased one, possibly suggesting that it learned to disregard $hw$ in
its predictions.
In the biased scenario, the network displayed higher robustness and
correctly learned all the monotonicity constraints. With the
increasing predictive performance, however, we can observe also a
drastic decrease in the demographic parity, indicating that the
network learned a discriminatory behavior from the data, even without
having access to the protected attribute $female$.

We remark that the probabilities computed with our procedure are
exact: aggregate results over multiple runs would only provide
information on the training stability.
Specialized techniques for the verification of single properties exist
(e.g. FairSquare~\cite{pfv-fair-fairsquare} for group fairness
verification), however, to the best of our knowledge no other
procedure is able to jointly verify all the properties reported above.

\section{Scalability}
\label{sec:scalability}

\customparagraph{Theoretical analysis.}
The previous Section provides a compelling argument in favor of
flexible verification routines, but scaling PFV beyond very
simple ML models is an open challenge.
By investigating the problem in very general terms, we hope to
facilitate the generalization of existing methods beyond their
original scope or the derivation of new approaches.
Notably, our work is not the first effort in adopting a unified
perspective on ML verification. Albeit restricted to robustness
verification of piecewise-linear neural networks, efforts in
developing a unified perspective led to significant practical
benefits~\cite{bunel2018unified}.

In order to identify effective scaling strategies, we first need to
pinpoint the sources of complexity in WMI-based PFV:
\begin{itemize}
\item[\textbf{S1}] is the number of convex polytopes to be integrated,
  i.e. the $\mu$s in \eqref{eq:wmidef}. Exactly enumerating these
  regions is at least has hard as solving \emph{model counting}
  (\#SAT), a \#P-complete problem~\cite{sang2005performing}. Even
  without accounting for the complexity of the enumeration, their
  sheer number can pose challenges to the verification. This factor is
  proportional to the combinatorial complexity of $\psys$ and $\ppro$,
  e.g. the number of activations in a NN.

\item[\textbf{S2}] is the complexity of integration, which is at least
  as hard as computing the volume of the polytope. Solving the volume
  computation exactly, or guaranteeing deterministic bounds on the
  approximation error are both NP-hard
  tasks~\cite{volumehardness}. This factor is mainly affected by the
  dimensionality of the joint $\prob{\bX, \bY}$.

\end{itemize}

Crucially, the problems originating from the PFV of ML systems fall
outside the restricted class of tractable WMI
computations~\cite{mpwmi,recoin}.
The outlook doesn't look very promising, as the factors affecting
\textbf{S1} and \textbf{S2} can quickly become unmanageable even for
relatively small problems.
Notice, however, that the LHS in \eqref{eq:pfvtask} doesn't have to be
computed exactly.
In fact, it can be possible to find a more tractable lower bound
$\tilde{P}$ such that:
\begin{align}
  k \le \tilde{P} \le \frac{\cprob{\propost,
      \propre}{\sys}}{\cprob{\propre}{\sys}} .
\end{align}
A common approach in the quantitative verification literature is
iteratively refining bounds $\tilde{P}_0 < \tilde{P}_1 < ... <
\tilde{P}_i$ until the threshold is matched, possibly ``sandwitching''
$k$ by tightening lower and upper bounds jointly.

In quantitative ML verification, FairSquare falls into this category,
where the under/over-approximations of $r_{parity}$~\eqref{eq:parity}
are unions of \emph{axis-aligned} hyperrectangles, which greatly
simplifies \textbf{S2}.
Similar under-approximations were recently adopted for scaling the
quantitative verification of ReLU NNs~\cite{preimageapprox},
restricted to $\ppro = \mathcal{U}$, with very promising results.
Generalizing these ideas to broader scenarios through our unified
framework is a promising research avenue that we are currently
investigating. \\

\customparagraph{A prototypical WMI-based verifier.}
While mitigating the hardness of integration (\textbf{S2}) is deferred
to future work, we substantiate the benefits of our framework in
addressing the combinatorial explosion of convex fragments of the
problem (\textbf{S1}).
To this extent, we devised a prototypical verifier, building on top of
the same exact WMI solver considered in Sec.~\ref{sec:motivating}.
This proof-of-concept algorithm, described in Alg.~\ref{alg:final}, is
meant to showcase (1) a generalized version of a pruning technique
used in deterministic verification, and (2) a simple incremental
verification strategy.

We turned to \emph{bound propagation} (BP), a common pruning technique
in ReLU network verification, with the goal of generalizing it to all
the classes of models described in Section~\ref{sec:mlmodelling}.
In its most basic form, BP forward-propagates upper and lower bounds
of each variable from input to output layers. Determining the
behavior of activations inside the relevant input interval, it is
then possible to simplify them with linear operations when they are
stable in $B$, i.e. always inactive ($\forall \bX \in B \:.\: \bW \bX
+ b \le 0$) or always active ($\forall \bX \in B \:.\: \bW \bX + b \ge
0$).
We observed that the dependency structure among variables of $\psys$,
which can be easily inferred for every class of model introduced so
far, can be used for propagating bounds $B$ to the relevant parts of
the system.
We then implemented a procedure $\alg{encodeSystem}(\psys, \encpre)$
that additionally accounts for the precondition, using it to compute
bounds to $\bX$ and propagate them during the encoding phase.
These bounds can be analytically computed in closed form for some
$\encpre$, such as $\epsilon$-balls around test points
\eqref{eq:localrob}, or they can be inferred via a linear number of
constrained optimization calls.
Then, whenever:
\begin{itemize}
  \item the behavior of $\sys$ would be encoded by a conditional
    clause $\ite(\Delta, \support_{T}, \support_{F})$, with $\Delta$
    being a condition on variables $\bX'$;
  \item $\bX' \in B \models \Delta$ (or $\bX' \in B \not\models
    \Delta$) can be checked efficiently, i.e. $\Delta$ is
    \emph{convex};
\end{itemize}
the procedure possibly simplifies the encoding to $\support_T$
($\support_F$ respectively). Other than ReLU NNs, conditions that are
easily checked include (possibly oblique) splits in tree-based models
and the conditions encoded in probability tables (or factors) of
graphical models.
In our empirical evaluation in Section~\ref{sec:experiments}, we
demonstrate how the benefits of BP also apply to the PFV of random forests.

Observing that most preconditions $\propre$ of interest are defined on
$\bX$ only, not involving $\sys$, the algorithm first computes $Z =
\cprob{\propre}{\sys} = \prob{\propre}$. Then, it partitions the
search space of $\cprob{\propost, \propre}{\sys}$ into an
\emph{ordered set} of disjoint sub-problems $\partition \defas [\phi_1,
  ... \phi_{\npartitions}]$, ensuring that $WMI(\Delta, w) =
\sum_{\phi \in \partition} WMI(\Delta \land \phi, w)$.
These are sequentially solved them until either $k \cdot Z$ is matched
by the partial result $\tilde{P}$ or the full search space is visited.
\begin{algorithm}
  \caption{ $\alg{WMIVerifier}(\pro, \psys, k)$} \label{alg:final}
  \begin{algorithmic}
    \STATE $\encpre, \encpost \leftarrow \alg{encodeProperty}(\propre, \propost)$
    \STATE $w_\pro \leftarrow \alg{encodePrior}(\ppro)$
    \STATE $w_{\sys} \leftarrow \alg{encodeSystem}(\psys, \encpre)$
    \STATE $Z \leftarrow \alg{WMI}(\encpre, w_\pro \cdot w_\sys)$
    \STATE $\partition \leftarrow \alg{Partition}(\npartitions, \encpre, \encpost, w_\pro, w_\sys)$
    \STATE $\tilde{P} \leftarrow 0$
    \FOR{$\phi \in \partition$}
    \STATE $\tilde{P} \leftarrow \tilde{P} + \alg{WMI}(\encpre \land \encpost \land \phi, w_\pro \cdot w_\sys)$
    \IF{$\tilde{P} > k \cdot Z$}
    \STATE \textbf{return} $true$
    \ENDIF
    \ENDFOR
    \STATE \textbf{return} $false$
  \end{algorithmic}
\end{algorithm}

While the worst-case complexity is the same of the naive approach
described in Sec.~\ref{sec:motivating}, using sensible selection and
ordering heuristics among partitions can lead to faster convergence.
In $\alg{Partition}(\npartitions, \encpre, \encpost, w_\pro, w_\sys)$,
a maximum of $\npartitions$ partitions are obtained by selecting a
subset of conditional statements $\ite(\Delta_1, \cdot, \cdot),
... \ite(\Delta_p, \cdot, \cdot)$ in $\support_\sys$ and fixing the
truth value of their conditions $\Delta_i$.
The criteria for scoring these conditions is based on approximating
$p(\Delta_i) \defas \sim \cprob{\Delta_i}{\encpre, \encpost, \sys}$
via sampling and fixing first the conditions minimizing $s(\Delta_i) =
|p(\Delta_i) - 0.5|$.
Intuitively, the heuristic aims at selecting the conditions that
split the search space more evenly with respect to the target
posterior.
A similar idea is adopted in ReLU NN verification~\cite{preimageapprox}
for selecting unstable activations to split on.
In our experiment, we set $\npartitions = 16$, selecting the
$log_2(\npartitions)$ conditions with the lowest score $s$.

In the following, we evaluate the effects of partitioning on the
verification runtime. Additionally, we compare the sampling-based
heuristic with random ordering.
The investigation of other forms of partitioning, such as those
splitting on input values rather than conditions induced by $\sys$, is
deferred to future work.

\section{Experiments}
\label{sec:experiments}
We evaluated the procedure described in Alg.~\ref{alg:final} on two
families of binary classifiers trained on synthetic data~\footnote{The
source code will be released upon acceptance.}.
In all experiments, $N$ input variables $\bX$ are independently
sampled from $\mathcal{N}(0,1)$, while the ground truth label is
computed as:
\begin{align*}
  c(\bX) \defas (\mathbf{M_1} \cdot \bX \cdot \mathbf{M_1}) \ge 0 ,
\end{align*}
where $\mathbf{M_1}$ and $\mathbf{M_2}$ are random matrices in $\{-1, 0,
1\}^{(N \times M)}$ and $\{-1, 1\}^{(M \times 1)}$ respectively, with
$M$ being a parameter controlling the complexity of the prediction
task (we fixed $M = 10$ in all experiments).
Being concerned with the mitigation of \textbf{S2}, we fixed the input
size $N = 3$, in order to reduce the integration time of each polytope
$\mu$.

As before, we encoded the prior $\ppro$ with a DET trained on a
separate dataset.
To provide further evidence of the flexibility of the approach, the
following experiments cover additional properties with respect to the
ones considered in Section~\ref{sec:motivating}, namely the
probabilistic variants of local robustness \eqref{eq:localrob} and
equivalence among predictors \eqref{eq:equivalence}.

We complemented ReLU NNs by additionally implementing random forests
(RFs) of DET-based binary classifiers. Our RFs leverage the DET
implementation and encoding procedures used for $\ppro$, training a
number of trees on different subsets of data to model conditionals
$\cprob{\bX}{y}$. The ensemble outputs the class maximizing these
conditional likelihoods:
\begin{align*}
  c(\bX) \iff \sum_i DET_i^\top(\bX) \ge \sum_i DET_i^\bot(\bX).
\end{align*}
where $DET_i^t(\bX)$ is trained on a subset $i$ of data to estimate
$P(\bX \:|\: y = t)$.

In both the benchmarks, we verified a ReLU NN having $3$ hidden layers
with $32$ activations each, which achieved an accuracy of $1.0$ on
the test set after training.

All experiments were performed on a machine with 28 Core (2.20GHz
frequency) and 128GB of RAM, running Ubuntu Linux 22.04.

\subsection{Probabilistic local robustness}
\label{sec:exp-rob}
Accounting for the distribution of inputs $\ppro(\bX)$, in this
experiment we check if the probability of changing prediction
in the $\epsilon$-ball around each test point is lower than $k = 0.1$.
This is in stark contrast with deterministic verification algorithms,
that would deem a test point non-robust even when a single
counterexample is found, regardless of its likelihood.
In practice, checking for non-robustness with $k = 0.1$, as opposed to
checking for $\epsilon$-robustness with $k = 0.9$, has the advantage
that fully robust instances, representing the large majority in most
settings, are more easily solved.
In this probabilistic setting, the verification of test instances has
three possible outcomes:
\begin{itemize}
  
  \item \emph{$\epsilon$-robust instances} (Rob), where no point in
    the $\epsilon$-ball violates the robustness property, i.e. $P =
    1$;
    
  \item \emph{probabilistically $\epsilon$-robust instances}
    ($\sim$Rob), where the probability of obtaining a different
    prediction in the $\epsilon$-ball is $P < 0.1$;
    
  \item \emph{non-robust instances} ($\neg$Rob), where the
    probability of a different prediction exceeded the threshold.
    
\end{itemize}

We run different experiments verifying the systems on $100$ test
points sampled from the ground truth distribution, with varying
$\epsilon$.
The RF considered in these experiments combine $20$ DETs and
achieved a test set accuracy of $0.85$ after training.

We first assessed the effectiveness of BP with $\epsilon \in [0.01,
  0.05, 0.1]$, reporting aggregated mean and standard deviation of
runtime across all test points (Table~\ref{tab:localrob_nopart}).
In both NN and RF verification, using the pruning strategy results in
a drastic runtime improvement.

We then evaluated the impact of the partition-based incremental
approach on harder instances, with $\epsilon \in [0.1, 0.15, 0.2]$.
We further evaluated the effect of the sampling-based heuristic by
considering a variant where partitions where sorted randomly.
Since the partitioning technique is only effective when the threshold
$k$ is exceeded, we report the runtimes of Rob, $\sim$Rob and $\neg$Rob
separately (Table~\ref{tab:localrob_part}).

As expected, easier instances, i.e. ones having smaller $\epsilon$,
are more likely to be probabilistically robust and are less affected
by the incremental approach.

For NNs, the advantage of the incremental strategy becomes evident on
the latter case as the hardness of the verification task increases, in
particular when $\epsilon = 0.2$. Importantly, the sampling-based
ordering heuristic improves upon random ordering.
Interestingly, the same improvement is not observed when verifying
RFs.
This might be due to the comparably smaller search space with respect
to verifying the NNs, or to the structural characteristics of our
DET-based ensemble.
In the following, we further investigate the matter in a verification
task involving \emph{both} NNs and RFs.

\begin{table}
  \centering
  
  \begin{tabular}{|c|c|c|c|}
  \hline
 & $\epsilon = 0.01$ & $\epsilon = 0.05$ & $\epsilon = 0.1$ \\ \hline\hline 

 \multicolumn{4}{|c|}{NN} \\ \hline
 No BP & \mstd{7.91}{18.26} & \mstd{83.85}{216.92} & \mstd{275.62}{635.56} \\
 BP & \mstd{0.35}{1.33} & \mstd{10.30}{39.07} & \mstd{71.44}{225.05} \\
\hline\hline

 \multicolumn{4}{|c|}{RF} \\ \hline
 No BP & \mstd{0.88}{3.09} & \mstd{31.80}{123.82} & \mstd{229.86}{775.49} \\
 BP & \mstd{0.17}{0.07} & \mstd{0.29}{0.58} & \mstd{1.35}{4.38} \\
\hline\hline

\end{tabular}

  \caption{
    \label{tab:localrob_nopart}
    Effect of BP on the runtime (in seconds) of Alg.~\ref{alg:final}
    on $\epsilon$-local robustness queries for both NNs and RFs.}
\end{table}

\begin{table}
  \centering
  
  \begin{tabular}{|c|ccc|}
 
 \hline & \multicolumn{3}{|c|}{$\epsilon = 0.1$}\\ 
  & Rob & $\sim$Rob & $\neg$Rob \\ 
 \hline \hline 

 NN & [87/100] & [5/100] & [8/100] \\ \hline
 - & \mstd{15.9}{23.0} & \mstd{153.8}{44.6} & \mstd{624.0}{532.0} \\
 Random & \mstd{16.5}{23.8} & \mstd{137.1}{39.5} & \mstd{394.4}{308.9} \\
 Sampling & \mstd{16.0}{23.0} & \mstd{150.7}{36.0} & \mstd{468.8}{367.5} \\
\hline
 RF & [80/100] & [11/100] & [9/100] \\ \hline
 - & \mstd{0.2}{0.0} & \mstd{3.3}{3.1} & \mstd{9.5}{10.8} \\
 Random & \mstd{0.2}{0.0} & \mstd{3.8}{2.7} & \mstd{8.7}{9.4} \\
 Sampling & \mstd{0.3}{0.1} & \mstd{36.8}{3.2} & \mstd{31.3}{8.1} \\
\hline
 \hline & \multicolumn{3}{|c|}{$\epsilon = 0.15$}\\ 
  & Rob & $\sim$Rob & $\neg$Rob \\ 
 \hline \hline 

 NN & [80/100] & [12/100] & [8/100] \\ \hline
 - & \mstd{49.6}{71.2} & \mstd{749.0}{384.7} & \mstd{2104.8}{1338.1} \\
 Random & \mstd{49.4}{70.9} & \mstd{612.7}{312.4} & \mstd{1240.2}{1024.4} \\
 Sampling & \mstd{49.4}{70.7} & \mstd{656.6}{313.1} & \mstd{999.6}{615.9} \\
\hline
 RF & [72/100] & [17/100] & [11/100] \\ \hline
 - & \mstd{0.2}{0.0} & \mstd{9.8}{14.8} & \mstd{50.9}{53.1} \\
 Random & \mstd{0.2}{0.0} & \mstd{9.6}{13.3} & \mstd{45.7}{47.8} \\
 Sampling & \mstd{0.3}{0.1} & \mstd{44.9}{15.1} & \mstd{69.6}{46.6} \\
\hline
 \hline & \multicolumn{3}{|c|}{$\epsilon = 0.2$}\\ 
  & Rob & $\sim$Rob & $\neg$Rob \\ 
 \hline \hline 

 NN & [72/100] & [16/100] & [12/100] \\ \hline
 - & \mstd{99.9}{144.6} & \mstd{1618.3}{815.5} & \mstd{5066.9}{3249.8} \\
 Random & \mstd{100.2}{145.0} & \mstd{1303.2}{682.3} & \mstd{3651.1}{2690.6} \\
 Sampling & \mstd{101.9}{147.3} & \mstd{1391.9}{836.4} & \mstd{2744.0}{1768.4} \\
\hline
 RF & [64/100] & [23/100] & [13/100] \\ \hline
 - & \mstd{0.2}{0.0} & \mstd{20.4}{33.9} & \mstd{134.7}{123.9} \\
 Random & \mstd{0.2}{0.0} & \mstd{19.4}{30.5} & \mstd{103.7}{90.8} \\
 Sampling & \mstd{0.4}{0.1} & \mstd{55.2}{34.6} & \mstd{126.6}{108.0} \\
\hline
\end{tabular}

  \caption{
    \label{tab:localrob_part}
    Effect of the partitioning-based incremental strategy on the
    harder instances of $\epsilon$-local robustness. Runtimes are
    reported separately for $\epsilon$-robust (Rob), proabilistically
    $\epsilon$-robust ($\sim$Rob) and non-robust ($\neg$Rob)
    instances.}
  
\end{table}

\subsection{Probabilistic equivalence of predictors}
\label{sec:exp-eq}
In this experiment, we turn to the task of verifying the equivalence
of two predictors with respect to a distribution of inputs
$\ppro(\bX)$.
For instance, this is useful for guaranteeing that a surrogate model
will behave similarly to a larger reference model, once deployed in an
uncertain environment.
%
%
Since we are comparing NNs with RFs, this task is out of reach of
model-specific verification algorithms.
The RF considered in these experiments is meant to be a surrogate
model that probabilistically behave similarly to the larger NN. We
downsized to $8$ DETs only, still achieving a test set accuracy
of $0.88$.

Similarly to the previous use case, we turn to the local variant
of~\eqref{eq:equivalence}, adopting the same precondition
of~\eqref{eq:localrob}.
Again, we evaluate our implementation on $100$ test instances,
reporting the runtime separately for:
\begin{itemize}
  \item \emph{$\epsilon$-equivalent instances} (Equ), where no point
    in the $\epsilon$-ball violates the equivalence property, i.e. $P
    = 1$;
  \item \emph{probabilistically $\epsilon$-equivalent instances}
    ($\sim$Equ), where the probability of the two models disagreeing
    in the $\epsilon$-ball is $P < 0.1$;
  \item \emph{non-equivalent instances} ($\neg$Equ), where the
    probability of disagreement exceeded the threshold.
\end{itemize}
Despite the reduction in size of the RF, the combination with the NN
results in a very challenging PFV task. We therefore considered the
smaller values of $\epsilon \in [0.01, 0.05, 0.1]$ while using BP.

As in the RF verification in the previous benchmark, the
sampling-based heuristic does not seem more effective than random
ordering, resulting effective in NN verification tasks only.
Nonetheless, the impact of the partitioning-based incremental strategy
is still significant, confirming its usefulness in mitigating the
explosion of the search space.

\begin{table}
  \centering
  
  \begin{tabular}{|c|ccc|}
 
 \hline & \multicolumn{3}{|c|}{$\epsilon = 0.01$}\\ 
  & Equ & $\sim$Equ & $\neg$Equ \\ 
 \hline \hline 

 NN vs. RF & [94/100] & [1/100] & [5/100] \\ \hline
 - & \mstd{0.1}{0.0} & \mstd{26.1}{0.0} & \mstd{54.1}{32.0} \\
 Random & \mstd{0.1}{0.0} & \mstd{26.0}{0.0} & \mstd{49.6}{36.2} \\
 Sampling & \mstd{0.1}{0.0} & \mstd{26.1}{0.0} & \mstd{53.7}{36.6} \\
\hline
 \hline & \multicolumn{3}{|c|}{$\epsilon = 0.05$}\\ 
  & Equ & $\sim$Equ & $\neg$Equ \\ 
 \hline \hline 

 NN vs. RF & [88/100] & [3/100] & [9/100] \\ \hline
 - & \mstd{1.5}{2.8} & \mstd{574.4}{295.0} & \mstd{406.7}{201.2} \\
 Random & \mstd{1.6}{2.8} & \mstd{585.1}{301.2} & \mstd{211.2}{86.4} \\
 Sampling & \mstd{1.5}{2.8} & \mstd{583.1}{297.3} & \mstd{212.9}{85.8} \\
\hline
 \hline & \multicolumn{3}{|c|}{$\epsilon = 0.1$}\\ 
  & Equ & $\sim$Equ & $\neg$Equ \\ 
 \hline \hline 

 NN vs. RF & [78/100] & [10/100] & [12/100] \\ \hline
 - & \mstd{14.4}{24.2} & \mstd{1631.4}{1391.2} & \mstd{4981.4}{4420.8} \\
 Random & \mstd{14.6}{24.6} & \mstd{1623.2}{1400.3} & \mstd{3100.1}{2531.0} \\
 Sampling & \mstd{14.6}{24.5} & \mstd{1639.0}{1405.1} & \mstd{3031.2}{2561.9} \\
\hline
\end{tabular}

  \caption{
    \label{tab:equivalence_part}
    Runtimes in $\epsilon$-local equivalence tasks. Runtimes are
    reported separately for $\epsilon$-equivalent (Equ),
    proabilistically $\epsilon$-equivalent ($\sim$Equ) and
    non-equivalent ($\neg$Equ) instances.}
  
\end{table}

\section{Conclusion and future work}
\label{sec:conclusion}

Our WMI-based perspective on the probabilistic verification of ML
systems has many benefits.
First, it enables flexible probabilistic reasoning over a broad range
of probabilistic models, including distributions over mixed
logical/numerical domains (\textbf{D1}).
These probabilistic models can be combined with many ML models of
practical interest and their combination (\textbf{D2}).
These aspects, paired with the ability to verify many different
properties (\textbf{D3}), including hyperproperties, offer
unprecedented flexibility in a single verification framework.
We believe that this unifying perspective can substantially contribute
to the systematization of this emerging field, and boost further
research on the topic.

Furthermore, we presented a theoretical analysis of the scalability of
the approach and implemented a prototypical $\alg{WMIVerifier}$, with
the goal of exploring two techniques aimed at reducing the
combinatorial explosion of the search space (\textbf{S1}).
We performed experiments on very challenging PFV tasks, including the
novel task of contrasting NNs and RFs.
We release both our solver and the benchmarks, providing a first
baseline for advancing the state of the art in WMI-based PFV.

In the following, we discuss its challenges and relevant research
directions.

\customparagraph{Scalability.}
Recent efforts have been devoted to scaling up WMI inference,
addressing the combinatorial explosion of complex weight
functions~\cite{spallitta2022smt,spallitta2024enhancing} or providing
statistical guarantees on approximations~\cite{abboud2022approximate}.
Combining these advances with the techniques developed in ML
verification is paramount to scalability.
While we moved the first steps in addressing \textbf{S1}, developing
approximations to the integrals is a necessary direction for tackling
\textbf{S2} and scale to high-dimensional problems, like those arising
in computer vision. In this regard, adopting incremental approaches
based on refining integral bounds, such as preimage
approximation~\cite{preimageapprox}, is a promising direction.


\customparagraph{Probabilistic neuro-symbolic verification.}
All the properties that we presented are defined on the \emph{concrete
input space} of the system under verification. Yet, being able to
define and verify properties using \emph{semantic} features is deemed
crucial step for advancing the trustworthiness of our AI
systems~\cite{towards-verified-ai}.
Since joint logical and probabilistic reasoning over $\sys$ and $\pro$
is achieved via a unified representation language and computational
tools, the classes of ML model that can be encoded in our framework
can appear in the definition of $\pro$. For instance, since
convolutional NNs with ReLU can be encoded, then $\pro$ can also be
defined in terms of these models, effectively implementing a
probabilistic variant of \emph{neuro-symbolic
verification}~\cite{xie2022neuro}.  Given a neural predicate that
detects stop signs in images, we could quantify the probability of
semantic properties like ``If a stop sign is in front of the camera, a
deceleration command is issued'' on systems operating at pixel level.

\customparagraph{Sequential systems.}
In this work, we focused on non-sequential systems. Yet, most ML
models are part of larger software and/or hardware systems with
memory.
Extending practical PFV tools like
PRISM~\cite{kwiatkowska2006quantitative} and
STORM~\cite{dehnert2017storm} with the concepts presented here would
significantly advance our ability to verify AI systems at large. From
a software development perspective, our framework can offer a unified
interface to many different approaches and tasks, simplifying their
integration into the existing verification software.


\section*{Acknowledgements}

  Funded by the European Union. Views and opinions expressed are
  however those of the author(s) only and do not necessarily reflect
  those of the European Union, the European Health and Digital
  Executive Agency (HaDEA) or The European Research Council. Neither
  the European Union nor the granting authority can be held
  responsible for them. Grant Agreements no. 101120763 - TANGO and
  no. 101110960 - PFV-4-PTAI.
  We acknowledge the support of the PNRR project FAIR - Future AI
  Research (PE00000013), under the NRRP MUR program funded by the
  NextGenerationEU.
  The work was partially supported by the project “AI@TN” funded by
  the Autonomous Province of Trento.

\bibliographystyle{plain}
\bibliography{paper}

\newpage

\onecolumn

\appendix

\section*{Use case: an income classifier}
\label{sec:app-motivating}

\subsection{Ground truth distribution and training}

\begin{table}[h]
  \centering
{\small
\begin{tabular}{|l|c|}
  \hline
  Female gender & $female \sim Bernoulli(0.5)$\\
  \hline
  Work hours per week & $hpw = 5 \cdot max(min(2, \mathcal{N}(7, 1)), 12)$ \\
  \hline
  Years of working experience & $yexp = max(min(0, \mathcal{N}(20, 7)), 55)$ \\
  \hline \hline
  Hourly wage & $hw = \mathcal{N}(1, 0.02) \cdot \frac{500 \cdot yexp + 30000}{1920}$ \\
  \hline
  Biased wage &  $hw_{b} = hw + hw \cdot N(01, 0.02) \cdot (1 - 2 \cdot \ind{feamale})$ \\
  \hline \hline
  Yearly income & $income = 48 \cdot hpw \cdot hw$ \\
  \hline
  Binary class & $y = (income > 35000)$ \\ \hline
\end{tabular}}
\caption{
  \label{tab:income}
  Parameters of the ground truth distribution adopted in the
  \emph{income} experiment.}
\end{table}

For these experiments, we borrowed the DET implementation and WMI
encoding of Morettin et al.~2020. The population models were trained
on $10000$ instances using hyperparameters $n_{min} = 1000$ and
$n_{max} = 2000$. These hyperparameters respectively dictate the
minimum and maximum number of instances per leaf.
The binary predictors were implemented in PyTorch~(Paszke et al., 2019)
as fully connected feedforward networks with two latent layers having
$8$ neurons each. The networks were trained on $10000$ labelled
instances minimizing the binary cross entropy loss with stochastic
gradient descent (batch size $200$, learning rate $0.005$ and no
momentum).
The predictive performance of the learned models was assessed on an
identically distributed test set with $1000$ instances. The WMI
problems resulting from the demographic parity, monotonicity and
robustness queries were solved using the exact solver
SA-WMI-PA (Spallitta et al., 2022).

\end{document}